\documentclass[preprint,12pt]{elsarticle}

\usepackage{amsmath}
\usepackage{amssymb}
\usepackage{algorithm}
\usepackage{algpseudocode}
\usepackage{booktabs}
\usepackage{multirow}
\usepackage{booktabs}
\usepackage{graphicx}
\usepackage{subcaption}
\usepackage[dvipsnames]{xcolor}
\usepackage{cancel}
\usepackage{ulem}
\usepackage[colorlinks=true, linkcolor=blue, citecolor=red, urlcolor=green]{hyperref}

\journal{Robotics and Autonomous Systems}

\begin{document}

\begin{frontmatter}

\title{Interpretable Responsibility Sharing as a Heuristic for Task and Motion Planning}

\author[1]{Arda Sarp Yenicesu\fnref{equal}}
\author[1]{Sepehr Nourmohammadi\fnref{equal}}
\author[1]{Berk Cicek}
\author[1]{Ozgur S. Oguz\corref{cor1}}

\cortext[cor1]{Corresponding author.\ead{ozgur.oguz@bilkent.edu.tr}}
\fntext[equal]{These authors contributed equally to this work.}
\affiliation[1]{organization={Department of Computer Engineering, Bilkent University},
            city={Ankara},
            postcode={06800}, 
            country={Türkiye}}

\begin{abstract}
This article introduces a novel heuristic for Task and Motion Planning (TAMP) named Interpretable Responsibility Sharing (IRS), which enhances planning efficiency in domestic robots by leveraging human-constructed environments and inherent biases. Utilizing auxiliary objects (e.g., trays and pitchers), which are commonly found in household settings, IRS systematically incorporates these elements to simplify and optimize task execution. The heuristic is rooted in the novel concept of Responsibility Sharing (RS), where auxiliary objects share the task's responsibility with the embodied agent, dividing complex tasks into manageable sub-problems. This division not only reflects human usage patterns but also aids robots in navigating and manipulating within human spaces more effectively. By integrating Optimized Rule Synthesis (ORS) for decision-making, IRS ensures that the use of auxiliary objects is both strategic and context-aware, thereby improving the interpretability and effectiveness of robotic planning. Experiments conducted across various household tasks demonstrate that IRS significantly outperforms traditional methods by reducing the effort required in task execution and enhancing the overall decision-making process. This approach not only aligns with human intuitive methods but also offers a scalable solution adaptable to diverse domestic environments. Code is available at \url{https://github.com/asyncs/IRS}.
\end{abstract}

\begin{keyword}
\begin{footnotesize}
Task and Motion Planning \sep Interpretable Robotics \sep Rule-based Learning
\end{footnotesize}
\end{keyword}

\end{frontmatter}

\section{Introduction}

\begin{figure}[h]
  \centering
  \includegraphics[trim={6cm 0cm 0cm 0cm}, clip, width=0.7\linewidth]{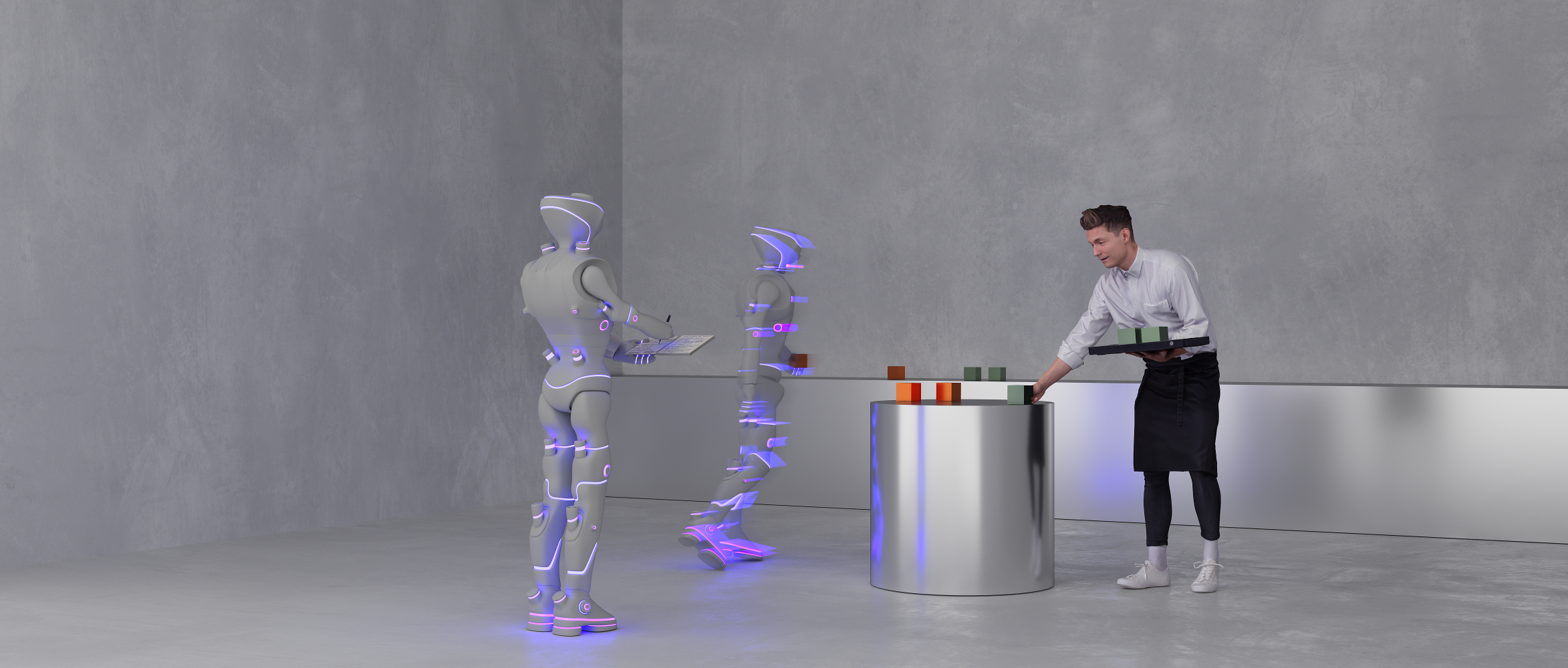}
  \caption{\textbf{The influence of Human-Centric Environmental Bias on TAMP.}} This study emphasizes the common human factor in both environment construction and task specifications present in household tasks (e.g., serving, cleaning, and caring). We highlight the role of Human-Centric Environmental Bias, represented through the use of auxiliary objects such as trays in kitchens, which are often created by humans for convenience. Our research proposes a systematic method to utilize these biases in an interpretable manner, aiming to enhance the efficiency and effectiveness of TAMP in human-designed environments.
  \label{fig:ga}
\end{figure}

Domestic robots are expected to handle household tasks defined by humans to improve their standards of living. These tasks include serving, cleaning, and caring in environments also created by humans. Due to this common human factor in sequential decision-making and manipulation problems, incorporating human-centric environmental bias into robots' decision-making processes can enhance their overall performance~\cite{qiao2022improving, he2017survey} (\textbf{Figure \ref{fig:ga}}).

In this study, we introduce the term \textbf{Human-Centric Environmental Bias} to describe the inherent structure embedded in domestic settings. When humans construct their environments, they introduce \textbf{auxiliary objects} (e.g., trays, pitchers) specifically designed to facilitate recurring tasks. Unlike approaches that leverage human bias through imitation learning from demonstrations, our perspective focuses on the inductive bias present in the environment itself. We propose that the mere presence of these human-designed objects encodes a preferred strategy (e.g., a tray implies a grouping strategy for serving). Our goal is to develop a systematic method to interpret and leverage these environmental cues, thereby achieving a more efficient TAMP formulation without requiring explicit human demonstrations.

Task and Motion Planning (TAMP) can effectively address these household tasks~\cite{dantam2016incremental, lagriffoul2016combining, hadfield2016sequential, ferrermestres2017combined, gaschler2018kaboum, shoukry2018smc, colledanchise2019towards}. Current TAMP research ~\cite{doi:10.1146/annurev-control-091420-084139, lgp, garrett2020pddlstream} often focuses on formulating these high-level tasks at a logical level, generating task plans using either uninformed or informed search, and finding feasible motions to execute these plans in a sequential or interleaved fashion.

However, these studies often overlook \textbf{human-centric environmental bias} due to the limitations of existing approaches. Uninformed task planners~\cite{pooja2016analyzing, correa2020resource} are domain-agnostic but suffer from exponential search spaces, limiting their applicability. Conversely, heuristic-based approaches~\cite{srivastava2013using, erdem2015integrating, youakim2020multirepresentation, braun2022rhhlgp} often rely on \textit{oracle} cost functions based on low-level physical metrics (e.g., displacement or energy), which fail to capture human preferences or the strategic utility of auxiliary objects. While recent deep learning methods~\cite{wang2021survey, deep_Affordance, mom, xian2023chaineddiffuser} show promise in adaptability by implicitly learning these biases from data, they lack transparency, making the resulting agents potentially unsafe for domestic deployment~\cite{Loveys2020, liu2023intention}.

Therefore, effective and interpretable task and motion planning require meeting two main criteria: \textbf{leveraging the human-centric environmental bias present in sequential decision-making and manipulation problems} and \textbf{enhancing interpretability in the decision-making process}.

\begin{figure}[t]
  \centering
  \includegraphics[trim={2.2cm 3.5cm 2.4cm 3.5cm}, clip,width=1\linewidth]{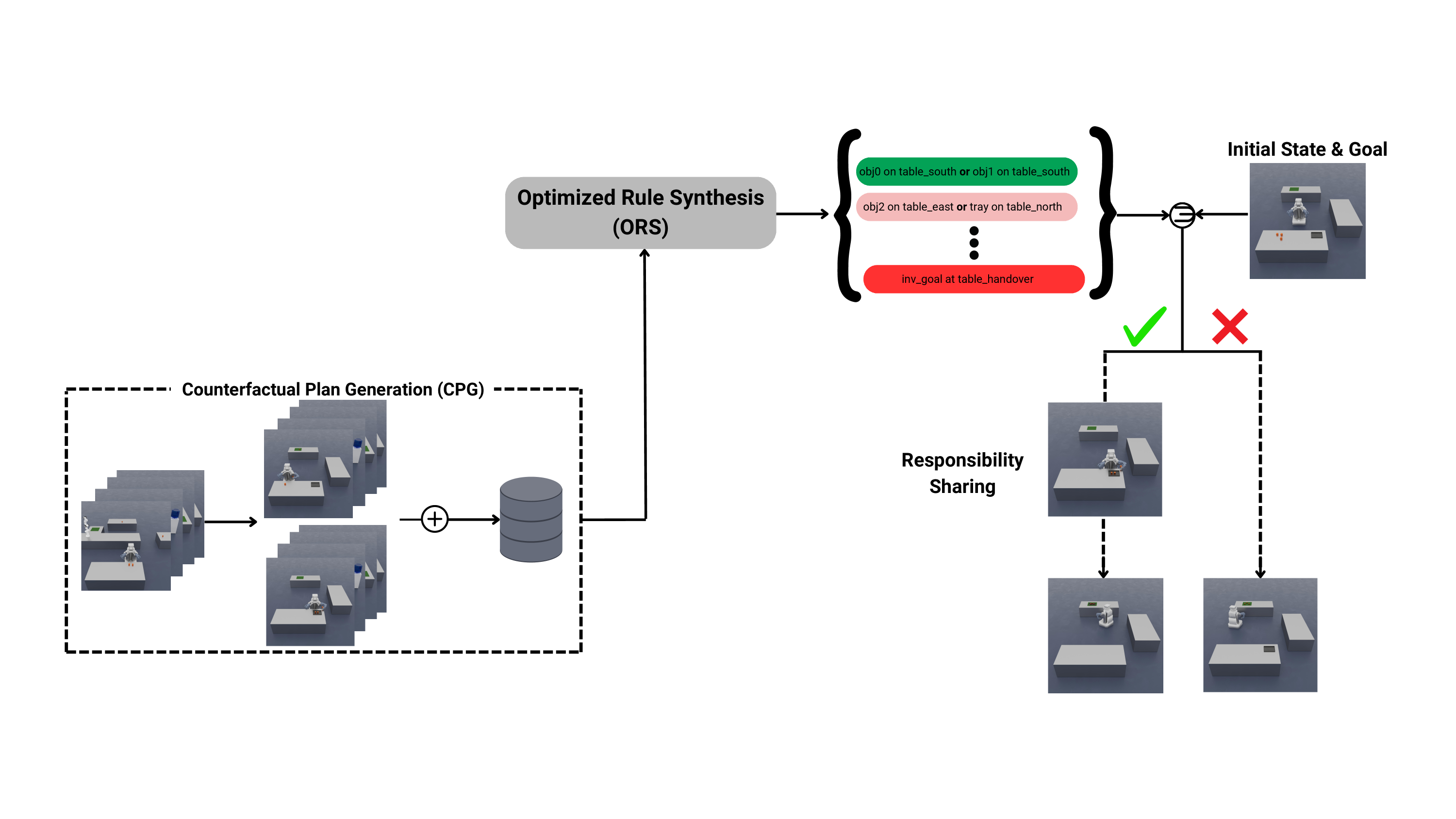}
  \caption{\textbf{The Interpretable Responsibility Sharing (IRS) Framework.} The pipeline consists of three core phases: (1) \textbf{Data Generation}, where Counterfactual Plan Generation (CPG) creates a labeled dataset based on physical effort; (2) \textbf{Rule Synthesis}, where Optimized Rule Synthesis (ORS) learns interpretable logical conditions for auxiliary object usage; and (3) \textbf{Heuristic Planning}, where IRS utilizes these rules to guide the TAMP solver by dynamically decomposing tasks into sub-problems.}
  \label{fig:irs}
\end{figure}

To meet these criteria, we propose a novel planning heuristic, \textbf{Interpretable Responsibility Sharing (IRS)}, which leverages human-centric environmental bias while providing an interpretable perspective on how an embodied agent uses this heuristic in its decision-making process (\textbf{Figure \ref{fig:irs}}). Crucially, IRS acts as a meta-heuristic that is orthogonal to existing TAMP solvers; it guides the search by dividing the problem into sub-problems independent of the underlying exploration strategy.

{Unlike traditional heuristics that compute low-level physical costs (e.g., displacement) online to guide the search, IRS utilizes these metrics strictly as offline supervisory signals during the learning phase. By \textit{compiling} physical costs into symbolic rules, our approach shifts the computational complexity of physical evaluation from inference time to the training phase, aligning with hierarchical sub-goal decomposition methods~\cite{HLGP}.

The core pipeline of our framework operates in three stages to bridge environmental bias and planning efficiency:
\begin{enumerate}
    \item \textbf{Data Generation (CPG):} We employ Counterfactual Plan Generation to create a dataset that labels whether using an auxiliary object improves performance for a given task configuration.
    \item \textbf{Rule Synthesis (ORS):} Using this dataset, Optimized Rule Synthesis learns interpretable logical rules that predict when auxiliary objects should be used.
    \item \textbf{Heuristic Planning (IRS):} During inference, IRS uses these rules to guide the TAMP solver. If the rules trigger \textbf{Responsibility Sharing (RS)}—the delegation of transport tasks to auxiliary objects—the planner decomposes the problem into efficient sub-goals; otherwise, it defaults to standard search.
\end{enumerate}

While the term \textbf{Responsibility Sharing} may evoke concepts from multi-agent systems~\cite{task_allocation}, we extend it here to the relationship between an agent and auxiliary objects. By delegating some responsibilities to auxiliary objects (e.g., placing the objects on a tray, which the agent then carries), the original problem is divided into sub-problems. This approach simplifies and enhances the effectiveness of the combined planning problem by leveraging inductive bias. It should be noted that this concept differs from tool usage, as using these objects is not mandatory to complete the task.

This study specifically aims to answer how embodied agents can systematically interpret and leverage these inductive biases, such as the strategic use of auxiliary objects, to enhance planning efficiency. Our primary focus is to develop an interpretable heuristic that identifies \textit{when} sharing responsibility is beneficial by compiling physical intuition into high-level logical rules. To isolate and validate this decision-making logic, we limit our experimental scope to fully observable settings and assume sufficient physical capacity for auxiliary tools. This abstraction allows us to evaluate the effectiveness of the proposed heuristic in capturing human-centric environmental bias without the confounding complexity of partial observability or dynamic bin-packing.

Overall, this work introduces the following contributions:
\begin{itemize}
    \item Interpretable Responsibility Sharing (IRS), a heuristic for task and motion planning, aiming to improve the effectiveness and interpretability of the agent by leveraging the inherent human-centric environmental bias.
    \item Optimized Rule Synthesis (ORS), a novel rule synthesis framework that integrates feature-based and relation-aware generators through a bounded greedy selection process. ORS introduces a \textit{Balance Score} objective to filter rules based on their \textit{informational sufficiency}, ensuring the synthesized policy captures the necessary physical context to ground robotic decision-making.
    \item A dataset investigating the benefits of using auxiliary objects in robotic scenarios, defined within a first-order logic framework, using Counterfactual Plan Generation (CPG).
\end{itemize}

We evaluate our IRS through three TAMP tasks suitable for domestic robots: serving, pouring, and handover. In the \textit{serving} task, objects are carried to their final destinations, while in the \textit{pouring} task, a source is distributed to objects at different locations using auxiliary objects like trays and pitchers. The \textit{handover} task offers a unique perspective on the boundaries of responsibility sharing, as a stationary robot acts as an auxiliary object, demonstrating the applicability of IRS in multi-robotic settings. A human experiment is conducted to observe the presence of Responsibility Sharing (RS) within human decision-making mechanisms. Our ablation studies provide valuable insights into decision-making performance and interpretability.
\section{Related Work}

\subsection{Task and Motion Planning (TAMP)}
Combined TAMP involves jointly solving high-level symbolic action planning and low-level motion planning ~\cite{doi:10.1146/annurev-control-091420-084139,ferrermestres2017combined,Integrating_Symbolic}. TAMP settings are defined by a combination of symbolic actions, states, and physical constraints, with tasks represented by either symbolic or physical goals. To identify feasible high-level action sequences for these goals, various TAMP planners have been developed. 

Early approaches often relied on linear, flow-like architectures~\cite{aSyMov, SMAP}. Modern solvers have evolved into complex, interleaved planners that generally fall into two distinct categories. Sampling-based methods, such as PDDLStream~\cite{garrett2020pddlstream} and FFRob~\cite{FFRob}, integrate symbolic planning with geometric reasoning by treating geometric constraints as black-box samplers; they generate candidate values for continuous parameters during the search and verify feasibility through external motion planning queries. In contrast, optimization-based strategies, most notably Logic-Geometric Programming (LGP)~\cite{lgp}, formulate the TAMP problem as a constrained mathematical program, optimizing symbolic choices and continuous trajectories simultaneously to satisfy physical constraints.

While these methods ensure physical consistency, TAMP inherently faces scalability challenges due to large action and state spaces, limiting the horizon of proposed plans and their adaptability to complex scenarios ~\cite{nair2019hierarchical, Driess_Long-Horizon}. Common strategies to address this include heuristic-guided search ~\cite{braun2022rhhlgp, Erdem2011CombiningHC,planner-independent} and breaking down the original problem into sub-problems ~\cite{HLGP,effort_level, hartmann2022long}.

\subsection{Rule-Based and Advanced Knowledge Graph Reasoning Methods}

\textbf{Rule-Based and Gradient Methods:}~Traditional rule-based models such as decision trees and rule sets, while transparent, often face limitations in scalability.  To enhance performance, ensemble methods like Random Forests have been employed, but their complexity often detracts from the interpretability required for robotics~\cite{hara2018making}. While Inductive Logic Programming (ILP)~\cite{muggleton1994inductive} offers verifiable logical rules, it struggles with scalability on noisy data. To address these limitations, gradient-based methods for discrete model training, such as the Straight-Through Estimator (STE) and Gradient Grafting~\cite{courbariaux2015binaryconnect, wang2020transparent}, have emerged. These techniques optimize discrete logical structures using continuous gradient information, bridging the gap between symbolic rigor and neural scalability.

\textbf{Knowledge Graph Reasoning:} In the domain of knowledge graphs, embedding techniques like TransE and RotateE~\cite{bordes2013translating, sun2019rotate} facilitate fast reasoning but suffer from opacity. In contrast, neural-symbolic methods blend the computational prowess of neural networks with the logical rigor of symbolic reasoning~\cite{rocktaschel2017end}, retaining interpretability through symbolic constraints.
Our approach with IRS leverages this synergy, combining gradient-based rule learning (RRL) with order-aware graph reasoning (CARL) to ensure both effectiveness and transparency.

\subsection{Interpretability in Robotics}
Interpretability in robotics is crucial as it ensures transparent and understandable decision-making processes, particularly in domestic and collaborative environments. Interpretable robotic systems are pivotal for enhancing trust, safety, and user acceptance, as they provide clear explanations for their actions and decisions~\cite{rodriguez2022towards,anjomshoae2019explainable,Paleja2023}. Various methods have been explored to foster interpretability in robotics, including rule-based approaches.

Among these, fuzzy rule-based methods utilize fuzzy logic to manage uncertainty and vagueness in robotic decision-making~\cite{samsudin2011highly,alonso2011hilk,mucientes2007quick}. Unlike fuzzy systems, where interpretability is often inversely proportional to the number of rules due to the ambiguity of overlapping membership functions, our approach leverages First-Order Logic (FOL) to define crisp, binary boundaries. While fuzzy interpretability suffers when the rule base becomes dense, interpretability in Task and Motion Planning benefits from \textit{descriptiveness}. Effective robotic planning requires rules that possess sufficient conditions to ground the action in physical reality (e.g., specifying locations or object counts)~\cite{tenorth2014knowledge}. Thus, distinct from fuzzy logic where \textit{less is more} prevents cognitive overload, we posit that for robotic agency, \textit{sufficient is more} is required to ensure causal validity~\cite{langley2019explainable}.

Alternatively, Graph Neural Networks (GNNs) offer a promising alternative for managing complex robotic manipulation tasks by utilizing relational data, which potentially enhances the interpretability of decisions~\cite{lin2022efficient,limeros2023towards,holzinger2021towards}. GNNs model task-relevant entities and their interrelations, supporting generalizable and somewhat interpretable robotic tasks. Nevertheless, the effectiveness of GNNs is often contingent upon the availability of substantial data sets, and their computational demands can be prohibitive, especially for real-time applications in robotics.
\section{Background}
Interpretable Responsibility Sharing (IRS) serves as a heuristic for Task and Motion Planning. In this study, we used Logic Geometric Programming (LGP) and Multi-Bound Tree Search (MBTS) to solve TAMP problems ~\cite{lgp, mbts}. During our dataset construction, we created counterfactual scenarios and used Individual Treatment Effect to determine the better scenario ~\cite{pearl2009causality}. Representative Rule-Based Learner (RRL) ~\cite{wang2021scalable} and Correlation and Order-Aware Rule Learning (CARL) ~\cite{he2023correlation} are the main components of ORS, providing interpretable responsibility sharing.

\subsection{Logic-Geometric Programming}
Logic Geometric Programming (LGP) serves as the foundational framework for solving TAMP problems. Let the configuration space $X \subset \mathbb{R}^n \times SE(3)^m$ represent $m$ rigid objects and $n$ articulated joints, starting from the initial condition $x_0$. The objective of LGP is to optimize a sequence of symbolic actions $a_{1:K}$, states $s_{1:K}$, and the associated continuous trajectory $x(t)$, where $t \in \mathbb{R}$ maps to $X$, in order to achieve a symbolic goal $g$. Positions, velocities, and accelerations are denoted by $\bar{x} = (x, \dot{x}, \ddot{x})$. The domain of symbolic states $s \in S$ and actions $a \in A(s)$ is discrete and finite, with state transitions $s_{k-1} \to s_k$ defined by a first-order logic language.
\begin{equation}
\begin{aligned}
& \hspace{10mm}\min_{x, s_{1:K}, a_{1:K}, K} \hspace{2mm}\int_0^{KT} c(\bar{x}(t), s_k(t)) \, dt \\
& \hspace{18mm}\text{s.t.}\hspace{1mm} \quad \bar{x}(0) = x_0, \\
& \hspace{7mm}\forall t \in [0, KT] :\hspace{2mm}h_{\text{path}}(\bar{x}(t), s_k(t)) = 0, \\
& \hspace{34mm} g_{\text{path}}(\bar{x}(t), s_k(t)) \leq 0, \\
& \forall k \in \{1, \ldots, K\} :\hspace{2mm}h_{\text{switch}}(\bar{x}(t_k), a_k) = 0, \\
& \hspace{34mm} g_{\text{switch}}(\bar{x}(t_k), a_k) \leq 0, \\
& \hspace{34mm}a_k \in A(s_{k-1}), \\
& \hspace{34mm}s_k \in \text{succ}(s_{k-1}, a_k), \\
& \hspace{34mm}s_K \in S_{\text{goal}}(g).
\end{aligned}
\label{lgp}
\end{equation}
At the heart of LGP lies the integration of discrete symbolic search (i.e., finding a plan skeleton) with simultaneous nonlinear optimization. This combined strategy computes trajectories that adhere to constraints and determines the feasibility of symbolic actions.

\subsection{Multi-Bound Tree Search}
\label{MBTS}
The discrete elements of the LGP formulation give rise to a decision tree populated with sequences of symbolic states. The feasibility of each pathway is assessed by solving the NLP produced by the state sequence. To mitigate the high computational cost of solving these NLPs, Multi-Bound Tree Search (MBTS) initially resolves relaxed versions of Equation \ref{lgp} to prune the search space ~\cite{mbts}. For this work, we utilize two bounds: $P_{seq}$ (joint optimization of mode-switches) and $P_{path}$ (full motion planning).

\subsection{Counterfactual Scenarios \& Individual Treatment Effect (ITE)}
We quantify the utility of auxiliary objects using the Potential Outcomes framework~\cite{pearl2009causality, 10.1145/3444944}. Let $W \in \{0, 1\}$ denote the binary treatment variable, where $W=1$ indicates the application of Responsibility Sharing (utilizing auxiliary objects) and $W=0$ indicates the baseline strategy (direct manipulation). Let $Y \in \mathbb{R}$ represent the outcome of interest, defined as the agent's performance cost (e.g., total displacement effort). For a specific task instance $i$, the Individual Treatment Effect (ITE) is defined as the difference between the potential outcomes under the treated and control conditions:
\begin{equation} 
\text{ITE}_i = Y_i(W = 1) - Y_i(W = 0)    
\label{ITE}
\end{equation}
In standard observational studies, one of these outcomes is typically unobserved (counterfactual). However, given the deterministic transition dynamics of our simulation, we can explicitly generate both potential outcomes for a fixed initial state $s_0$. We generate the factual plan $P$ (standard search, $W=0$) and the counterfactual plan $P'$ (constrained sub-goal search, $W=1$) to directly compute $\text{ITE}_i$. This value determines the ground-truth label for training: a negative treatment effect ($\text{ITE}_i < 0$) implies a reduction in cost, indicating that the use of the auxiliary object yields a positive performance gain.

\subsection{Rule-Based Representation Learner (RRL)}\label{RRL}
To extract interpretable logical conditions from high-dimensional continuous data, we employ the Rule-Based Representation Learner (RRL)~\cite{wang2021scalable}. Traditional decision trees often struggle with scalability on large continuous datasets. RRL addresses this by merging the interpretability of discrete rules with the efficiency of gradient-based optimization. The model utilizes a binarization layer to discretize continuous features and employs Gradient Grafting to bridge the gap between discrete logic and continuous optimization. The update rule is defined as:
\begin{equation}
    \hat{g} = g_{\text{continuous}} \times I(\sigma_{\text{logical}}(x) \geq 0.5) + g_{\text{discrete}} \times (1 - I(\sigma_{\text{logical}}(x) \geq 0.5))
\end{equation}
This mechanism allows RRL to leverage scalable gradient descent for training while outputting the crisp, discrete logical rules required for symbolic planning.

\subsection{Correlation and Order-Aware Rule Learning (CARL)}
\label{sec:CARL}
While RRL efficiently processes flat feature vectors, TAMP environments are inherently relational (e.g., spatial topologies). To capture these dependencies, we utilize Correlation and Order-Aware Rule Learning (CARL)~\cite{he2023correlation}. Unlike traditional methods that treat rule bodies as unordered sets of atoms, CARL explicitly models the \textit{semantic order} of relations within a path to predict the most likely rule head.

To distinguish the semantic role of a relation based on its position in the rule chain (e.g., first vs. last condition), the model augments relation embeddings with sinusoidal positional encodings. These encoded inputs are then processed via attention mechanisms to enforce semantic consistency.

Functionally, CARL employs the attention mechanism in two complementary phases to turn symbolic sequences into rich semantic representations.

First, in the \textbf{Correlation Module}, a multi-head self-attention layer is applied over all relations to estimate a global correlation matrix. An \textit{active-relations selector} uses this to choose only the most informative relations as queries. This prunes the search space, reducing the complexity of relation interactions and focusing resources on relevant candidates.

Second, in the \textbf{Semantic Learner}, multi-head cross-attention is used to recursively merge adjacent relations in the rule body. Here, the query $Q$ represents the current rule-body embedding, while the keys $K$ and values $V$ represent candidate relations. The attention weights effectively determine which relations are semantically compatible with the current rule context, compressing the rule body while enforcing logical consistency.

The model minimizes the cross-entropy loss over the set of training triples $Z$ to maximize the likelihood of the correct rule head $r_h$ given the rule body~$r_b$:
\begin{equation}
\mathcal{L} = - \sum_{(r_b, r_h) \in Z} \sum_{k=0}^{|R|} v_{r_hk} \log \theta_{zk}
\end{equation}
where $v_{r_h}$ is the one-hot ground truth vector for the head relation. The term $\theta_z \in \mathbb{R}^{|R|+1}$ represents the predicted probability distribution over possible rule heads (including a null head) derived from the attention outputs. Consequently, $\theta_z$ serves as the confidence score during rule extraction, ensuring that selected rules are both logically valid and semantically consistent with the underlying knowledge graph.
\section{Methodology}
We aim to leverage human-centric environmental bias in sequential decision-making and manipulation problems without compromising interpretability. To achieve this, we formulate \textbf{Responsibility Sharing (RS)} as a mechanism where auxiliary objects share the task load with the agent—for instance, using a bag to hold objects during transport rather than manipulating them directly. The following subsections detail the mathematical formulation of the problem (Section \ref{problem}) and the three core components of our framework: dataset generation (CPG), rule synthesis (ORS), and the heuristic planner (IRS).

\subsection{Problem Setup}\label{problem}
We consider fully observable domains with a symbolic state space defined under first-order logic $S$, configuration space $X$, action space $A$, and deterministic transition function $\delta : S \times A \rightarrow S$. We assume a finite set of goals $G$. Each $g \in G$ is a binary condition function $g : S \rightarrow \{0,1\}$ indicating if a symbolic state is a goal state $S_{\text{goal}}(g)$ or $s_g$. The goal is to reach a goal state $s_g$ starting from an initial state $s_0$ following a task plan $P = \{a_1,a_2,...,a_T\}$, where $T$ is the required number of actions.

Following prior TAMP formulation ~\cite{doi:10.1146/annurev-control-091420-084139, lgp, garrett2020pddlstream, toussaint2018differentiable}, we employ a search among the symbolic state space $S$. While this search is typically guided by heuristics, IRS acts as an \textbf{orthogonal meta-heuristic}: it guides the search by structurally dividing the problem into sub-problems, independent of the underlying exploration strategy. In this study, we utilize a \textbf{logically unguided search} to isolate and validate the performance gains provided specifically by the IRS formulation.

In our formulation, we adopt the concept of responsibility sharing, where the agent can utilize auxiliary and non-mandatory objects if inferred to increase effectiveness. To efficiently handle the combinatorial complexity of this choice, we decompose the TAMP problem into a set of specialized sub-problems, $\Psi$. This design choice allows us to isolate the high-level decision (whether to use a tray) from the low-level motion planning, treating Responsibility Sharing as a distinct meta-heuristic. The original problem is divided into sub-problems $\Psi$, which define the usage of the auxiliary objects (e.g., first carry objects on the tray, then carry the tray to the target position). These sub-problems are defined under the set of available actions, $A_\psi \subset A$, and the set of sub-goal states, $S_{\text{goal}}(g_\psi)$. Hence, the original objective remains, but with additional requirements to satisfy each sub-goal state where auxiliary objects are utilized.

\subsection{Counterfactual Plan Generation}
To train our rule learner, we require a dataset that objectively quantifies the physical benefit of using auxiliary objects. Since manual human labeling is costly and subjective, we introduce Counterfactual Plan Generation (CPG) to automatically extract ground-truth labels by systematically comparing alternative execution strategies. CPG aims to generate sequential task plans that utilize the auxiliary objects available in the environment (\textbf{Figure \ref{fig:cpg}}). This method involves three main steps:

\begin{figure}[t]
  \centering
  \includegraphics[trim={5cm 6.2cm 4.8cm 6.2cm}, clip,width=1\linewidth]{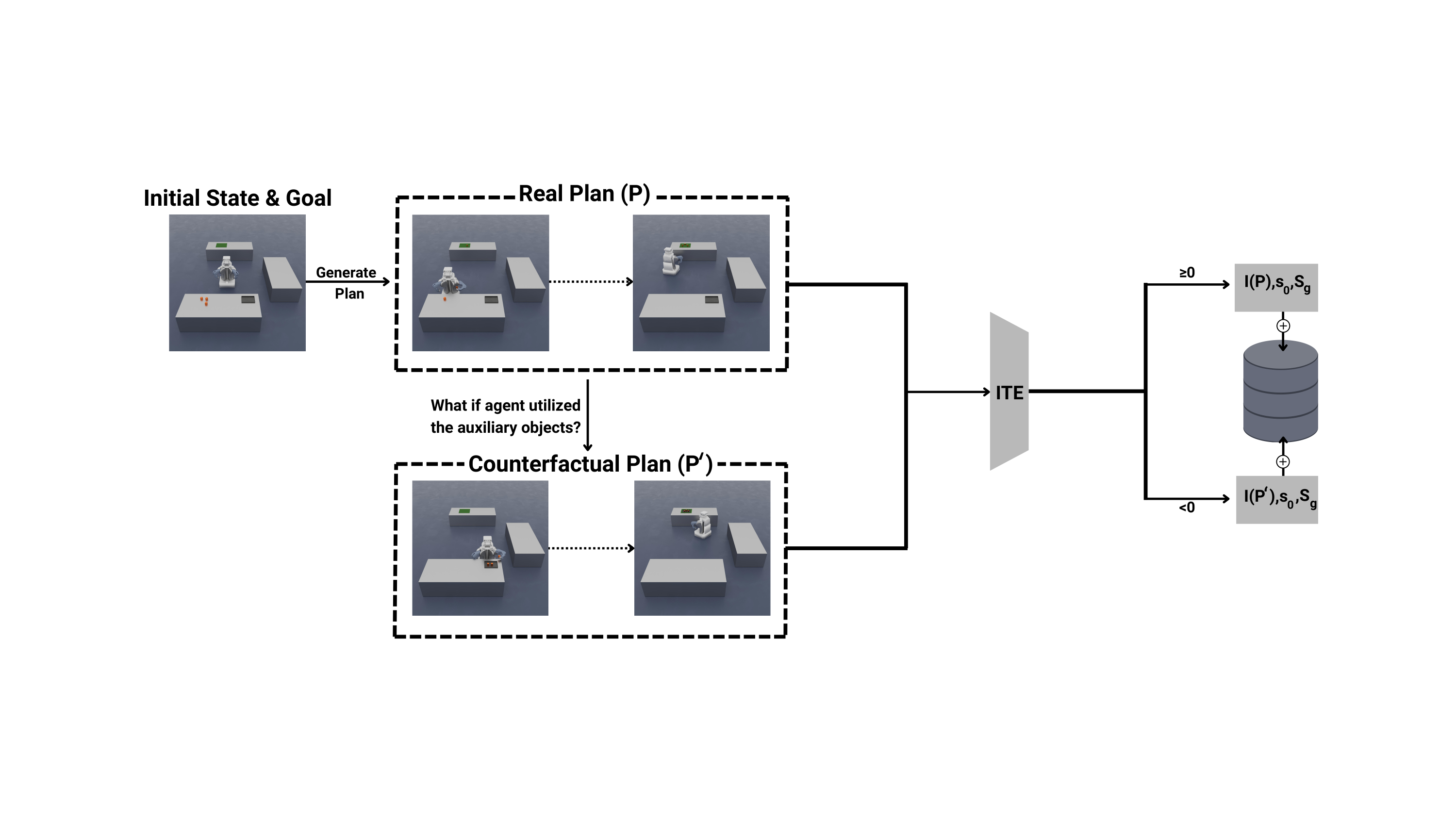}
  \caption{\textbf{Counterfactual Plan Generation (CPG) for Dataset Construction.} The process compares a standard Real Plan ($P$) against a Counterfactual Plan ($P'$) that specifically enforces the use of auxiliary objects. By calculating the Individual Treatment Effect (ITE) based on the cost difference between $P$ and $P'$, we determine the ground-truth label for each scenario (Positive if auxiliary usage reduces cost, Negative otherwise), resulting in the labeled dataset $D$ used to train ORS.}
  \label{fig:cpg}
\end{figure}

\subsubsection{Generating a Feasible Plan}
The first step in calculating the treatment effect is to establish a baseline performance metric (the control group). We achieve this by generating a \textit{feasible plan} using standard uninformed search, which represents the agent's default behavior when it does not explicitly prioritize responsibility sharing. Sequential decision-making and manipulation problems are defined by logical states, actions, goals, and physical constraints. Following the TAMP formulation, the agent employs a search in the state space following a transition function, $s'=\delta(s,a)$. Starting from the initial state $s_0$, the search algorithm expands from this state by taking actions $a_t$, transitioning to the next state $s'$, and gradually reaching a goal state $s_g$. This generates a series of actions resulting in a task plan, $P = \{a_1,a_2,...,a_T\}$. CPG utilizes Multi-Bound Tree Search ~\cite{mbts} as the search algorithm for finding a task plan due to its domain-agnostic and geometrically aware nature.

\subsubsection{Generating a Counterfactual Plan}\label{counterfactual_plan}
Next, we generate a counterfactual plan using the real plan that was generated in the previous step. The aim of this counterfactual generation is to determine how the plan would change if we utilized the auxiliary objects. Since the real plan is generated by the uninformed search, it does not utilize these objects as they increase the depth of the solution. Using the same initial state $s_0$, transition function $\delta$, and search method (which is uninformed), we introduce an additional sub-goal $s_i$, that satisfies the responsibility sharing conditions (e.g., if a tray will be used to carry the objects, they are first placed on the tray). 
The search expands from the initial state $s_0$ until it reaches the goal state $s_g$ while satisfying the intermediate state $s_i$. This generates a counterfactual series of actions, forming a counterfactual task plan, $P' = \{a_1,a_2,...,a_{T'}\}$.

\subsubsection{Dataset Construction}\label{dataset}
By repeating the first two steps, we generate multiple real and counterfactual plan pairs, $P$ and $P'$, based on the initial and goal states, $s_0$ and $s_g$. To decide which plan yields better performance, we employed Individual Treatment Effect (ITE) formulated as Equation \ref{ITE}. We selected the $L^2$ norm as the measure for the first two tasks, \textit{serving} and \textit{pouring}, and Manhattan distance for the last task, \textit{handover}, due to obstacles in the environment. 
After selecting the effective plan $P^*$ among the pairs, we use it as a binary label $l$ for the initial and goal state pair: positive if utilizing auxiliary objects yields better performance and negative otherwise.
We compile them into a dataset, $D = \{(s_o^1, s_g^1, l^1),...,(s_o^n, s_g^n, l^n)\}$, to train ORS, where $n$ represents the number of problems generated.

To ensure the validity of this labeling process and rule out potential skewness caused by the uninformed search (which prioritizes plan depth), we conducted a validation study on the serving task (detailed in Section 5.1). We compared our dataset labels against a planner-independent geometric heuristic that calculates the optimal $L^2$ path cost for both direct (individual transport) and auxiliary-based (shared responsibility) strategies. The results showed a strong alignment between our planner-derived labels and the geometric ground truth. This confirms that our dataset accurately reflects the \textbf{intrinsic importance of object usage} based on environmental structure, rather than exploiting the disadvantages of the uninformed search algorithm.

\begin{figure}[t]
  \centering
  \includegraphics[trim={1.9cm 11cm 1.9cm 11cm}, clip,width=1\linewidth]{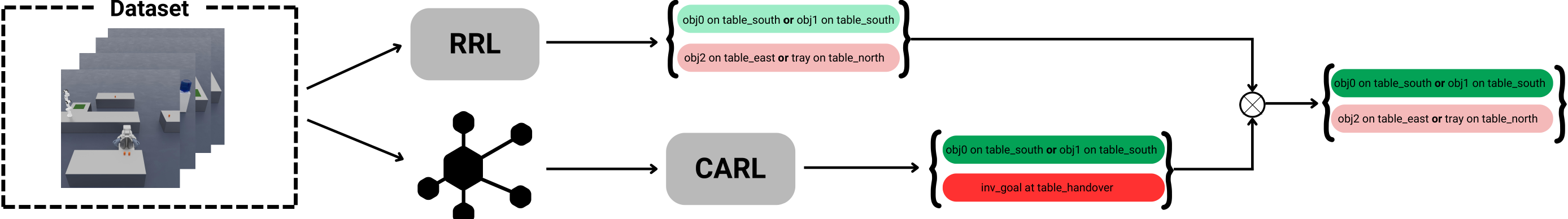}
  \caption{\textbf{The Optimized Rule Synthesis (ORS).} The framework takes the labeled dataset $D$ as input and employs two complementary generators: RRL (extracting structural logic) and CARL (extracting relational dependencies). The resulting candidate rules are pooled and iteratively selected via a bounded greedy search that optimizes a \textit{Balance Score}, weighing accuracy against interpretability. The final output is a stable, concise set of conditions ($S_c$) that dictates when Responsibility Sharing should be applied.}
  \label{fig:ors}
\end{figure}

\subsection{Optimized Rule Synthesis (ORS)}
\label{sepehr_method}

Optimized Rule Synthesis (ORS) is designed to learn interpretable rules that specify precisely when the agent should employ Responsibility Sharing (RS) with auxiliary objects. The method operates on a supervised dataset $D = \{(x_i,y_i)\}_{i=1}^N$, constructed via Counterfactual Plan Generation (CPG). Here, $x_i = \Phi(s_0^i,g^i) \in \mathbb{R}^d$ represents the feature encoding of the $i$-th initial–goal state pair, and $y_i \in \{-1,+1\}$ indicates the ground-truth label: $+1$ if using RS is beneficial, and $-1$ otherwise.

The feature encoding function $\Phi(s, g)$ maps the symbolic state $s$ and logical goal $g$ into a fixed-size binary feature vector $x \in \{0, 1\}^d$. This vector consists of flattened ground predicates representing the environment's configuration and task constraints. Specifically, the feature space $\Phi$ encompasses:
\begin{itemize}
    \item \textbf{Spatial Configuration:} Boolean indicators for the position of all manipulable and static objects relative to discrete locations (e.g., $obj_i\_on\_table_{north}$, $tray\_on\_table_{south}$).
    \item \textbf{Goal Constraints:} Boolean flags specifying the target conditions for the task (e.g., $goal\_at\_table_{east}$).
    \item \textbf{Action \& Context Primitives:} Indicators for environmental capabilities (e.g., $is\_helper\_exist$) and the set of high-level abstract actions valid for the current task domain (e.g., $action_{pour}$, $action_{handover}$, $action_{pick}$).
\end{itemize}
For RRL, this binary vector is used directly as input. For CARL, these active predicates are projected into a knowledge graph structure, where entities (e.g., objects, tables) are nodes and their spatial or functional relationships serve as edges.

To ensure robustness and coverage, ORS acts as an ensemble selector. It integrates two distinct rule-learning architectures as complementary generators: \textbf{Rule-based Representation Learner (RRL)}~\cite{wang2021scalable}, which constructs logical rules from discretized continuous features, and \textbf{Correlation and Order-Aware Rule Learning (CARL)}~\cite{he2023correlation}, which extracts rules from a knowledge graph, preserving correlation and order information. Rather than using these models directly, ORS aggregates their outputs and selects a sparse, high-performing subset of rules that balances accuracy with interpretability.

\subsubsection{Joint Rule Pool Generation}

To achieve a robust rule set, ORS employs a dual-generator strategy that leverages the distinct strengths of two architectures. We utilize RRL to capture strict structural constraints (e.g., explicit negations and specific object placements) from the feature vectors, while CARL is employed to extract complex relational dependencies and semantic chains from the knowledge graph. In the initial phase, both RRL and CARL are employed to generate a comprehensive set of candidate rules from the dataset $D$. Let $\mathcal{R}^{\text{CARL}}$ and $\mathcal{R}^{\text{RRL}}$ denote the sets of rules extracted by CARL and RRL, respectively. These are combined into a joint rule pool $\mathcal{R} = \mathcal{R}^{\text{CARL}} \cup \mathcal{R}^{\text{RRL}}$, which is subsequently deduplicated to remove semantically equivalent rules sharing the same antecedent and label.

Each candidate rule $r \in \mathcal{R}$ is defined by two components: a Boolean condition $\varphi_r: \mathbb{R}^d \to \{0,1\}$ (e.g., tray is present $\land$ object count $\geq 3$) evaluated on input $x_i$, and a label $\ell(r) \in \{-1,+1\}$ indicating the recommended decision when the condition holds.

\subsubsection{Rule-Set Classification}

ORS functions as a meta-learner that selects a subset $S \subseteq \mathcal{R}$ to form an interpretable classifier. This classifier maps feature vectors $x$ to RS decisions $\hat{y}$ via a voting mechanism. For a single rule $r$, the prediction on instance $x_i$ is determined by its condition satisfaction:
\begin{equation}
    \hat{y}_i^{(r)} =
    \begin{cases}
    \ell(r), & \text{if } \varphi_r(x_i)=1,\\
    -\ell(r), & \text{if } \varphi_r(x_i)=0.
    \end{cases}
\end{equation}
This formulation implies that a rule predicts its assigned label if satisfied, and the opposite label otherwise. The agreement of a rule with the ground truth is given by the indicator $\gamma_i(r) = \mathbb{I}[\hat{y}_i^{(r)} = y_i]$, and the individual accuracy of a rule $r$ over the dataset is defined as $\mathrm{Acc}(r;D) = \frac{1}{N}\sum_{i=1}^N \gamma_i(r)$.

To aggregate predictions from the selected subset $S$, we employ a signed voting scheme. The ensemble score is calculated as $s_S(x) = \sum_{r\in S} \varphi_r(x)\,\ell(r)$. The final classification decision $h_S(x)$ is then derived by thresholding this sum:
\begin{equation}
    h_S(x) =
    \begin{cases}
    +1, & \text{if } s_S(x) > 0,\\
    -1, & \text{otherwise}.
    \end{cases}
\end{equation}
The performance of the rule set is evaluated on a validation split $D_{\mathrm{val}}$ using the standard accuracy metric $\mathrm{Acc}(S;D_{\mathrm{val}}) = \frac{1}{|D_{\mathrm{val}}|}\sum_{i} \mathbb{I}[h_S(x_i) = y_i]$.

\subsubsection{Interpretability and Balanced Score}

In the domain of Task and Motion Planning (TAMP), we define interpretability not as sparsity (minimizing conditions), but as \textit{informational sufficiency} or \textit{descriptiveness}. While general machine learning often prefers shorter rules to prevent overfitting, robotic decision-making requires rules that capture sufficient physical context to justify an action to a human observer and grounded task execution~\cite{tenorth2014knowledge,langley2019explainable}.

For instance, a sparse rule such as \textit{Positive} $\longleftarrow$ $obj_{south}$ recommends an action based solely on object location, failing to explain the physical mechanism enabling the task. In contrast, a longer, context-aware rule such as \textit{Positive} $\longleftarrow$ $obj_{south}$ $\land$ \textit{helper exists} $\land$ $\neg goal_{south}$ explicitly identifies the necessary preconditions (e.g., the existence of a tray/helper) that make the action valid. To quantify this preference for context, we define the interpretability of a rule set $S$, denoted $\mathcal{I}(S)$, as the average rule length normalized by the maximum possible conditions $L_{\max}$:
\begin{equation}
    \mathcal{I}(S) = \frac{1}{|S|} \sum_{r \in S} \frac{\text{length}(r)}{L_{\max}}
\end{equation}
where $\text{length}(r)$ is the number of atomic conditions in the rule antecedent.

However, maximizing length arbitrarily can obscure the causal explanation through the \textit{dilution effect}~\cite{miller2019explanation}. To distinguish between necessary causal context and arbitrary or delusional verbosity, we utilize a Balance Score that jointly optimizes Accuracy and Interpretability:
\begin{equation}
    \mathrm{Score}(S; D_{\mathrm{val}}) = \alpha \cdot \mathrm{Acc}(S; D_{\mathrm{val}}) + (1-\alpha) \cdot \mathcal{I}(S)
    \label{eq:balance_score}
\end{equation}
By maximizing this combined objective, the system favors rules that are sufficiently descriptive (high $\mathcal{I}(S)$) to be intelligible. Crucially, the accuracy term ($\mathrm{Acc}$) acts as a semantic filter: conditions that increase length without contributing to (or maintaining) the predictive validity of the rule are rejected, ensuring that the retained predicates represent meaningful physical constraints rather than noise.

\subsubsection{Iterative Search Procedure}

Finding the optimal subset of rules that maximizes the Balanced Score is a combinatorial problem. Since enumerating all subsets is intractable for a realistic rule pool size $|\mathcal{R}|$, ORS employs a bounded greedy search strategy. This process is controlled by two hyperparameters: $\beta$, the number of random repetitions, and $K_{\max}$, the maximum number of rules allowed in set $S$.

To ensure generalization and reduce variance, the search procedure is repeated $\beta$ times (with $\beta=10$ in our experiments). In each iteration, the dataset $D$ is randomly partitioned into training ($D_{\mathrm{train}}$), validation ($D_{\mathrm{val}}$), and test ($D_{\mathrm{test}}$) subsets. The candidate generators, CARL and RRL, are trained exclusively on $D_{\mathrm{train}}$ to populate the rule pool $\mathcal{R}^{(\mathrm{iter})}$.

The selection process begins with an empty set $S = \emptyset$. In each step, the algorithm greedily adds the rule $r \in \mathcal{R}^{(\mathrm{iter})} \setminus S$ that yields the largest increase in $\mathrm{Score}(S; D_{\mathrm{val}})$. This strictly relies on the validation set to guide selection, preventing data leakage from the test set. The process terminates when no remaining rule improves the score or when $|S|$ reaches $K_{\max}$. After $\beta$ repetitions, the rules selected most frequently across all runs are aggregated to form the final, stable condition set $S_c$. The complete procedure is detailed in Algorithm~\ref{ORS}.

\begin{algorithm}[h]
\begin{footnotesize}
\caption{Optimized Rule Synthesis (ORS)}
\begin{algorithmic}[1]
    \State \textbf{Input:} Dataset $D = \{(x_i,y_i)\}_{i=1}^N$
    \State \textbf{Hyperparameters:} $\beta$ (repetitions), $K_{\max}$ (max rules), $\alpha$
    \State \textbf{Output:} Final RS rule set $S_c$

    \State $OptimalRuleSets \gets \emptyset$
    \For{$iteration = 1$ \textbf{to} $\beta$}
        \State $(D_{\text{train}}, D_{\text{val}}, D_{\text{test}}) \gets \text{SplitData}(D)$
        \State $AllRuleConditions \gets \text{CARL}(D_{\text{train}}) \cup \text{RRL}(D_{\text{train}})$
        \State $S \gets \emptyset$
        \Repeat
            \State Choose $r\in AllRuleConditions \setminus S$ that maximizes $\mathrm{Score}(S\cup\{r\};D_{\text{val}})$
            \If{$\mathrm{Score}(S\cup\{r\};D_{\text{val}}) > \mathrm{Score}(S;D_{\text{val}})$ \textbf{and} $|S|<K_{\max}$}
                \State $S \gets S \cup \{r\}$ \Comment{Generating Rules}
            \Else
                \State \textbf{break}
            \EndIf
        \Until{no improvement}
        \State Append $S$ to $OptimalRuleSets$
    \EndFor
    \State $S_c \gets \text{Aggregate}(OptimalRuleSets)$ \Comment{Keep rules selected frequently across runs}
    \State \Return $S_c$
\end{algorithmic}
\label{ORS}
\end{footnotesize}
\end{algorithm}

\subsection{Interpretable Responsibility Sharing as a Heuristic}

ORS generates logical rules that dictate the appropriate conditions under which the agent should share responsibilities with auxiliary objects. If the agent decides to use the auxiliary objects, sub-goals are created to satisfy the necessary conditions for their usage. These conditions are object-dependent and naturally occur during their usage. For example, if the agent wants to use a tray to carry objects, the objects need to be placed on the tray first. These preconditions create clear and intuitive sub-goals for the agent to operate effectively.

When the ORS conditions are satisfied ($S_c$ is true for $s_0, s_g$), the agent adopts the IRS strategy. By introducing a sub-goal-based formulation, our approach increases the number of generated plans compared to conventional MBTS if we consider each sequence from a state to a sub-goal as a different plan. However, this naive approach is computationally expensive for a combined task and motion planning approach since the solution time is highly dominated by the geometric calls made by the LGP approach, and these calls should be minimized for efficiency ~\cite{braun2022rhhlgp}.

Solving these sub-problems sequentially with full geometric validation is computationally prohibitive due to the high cost of non-linear optimization. To ensure tractability, we utilize the bound-deferral strategy of Multi-Bound Tree Search (MBTS). We prioritize $P_{\textrm{seq}}$ for optimizing mode-switch sequences (e.g., changing from free space movement to object grasping) and defer $P_{\textrm{path}}$, the complete problem (i.e., considering kinematics, dynamics, and including obstacle and collision checks), until its full realization. Given this abundance, we avoid solving joint motion separately multiple times and instead utilize the found mode-switch sequences for solving a complete motion plan. This necessitates modifications to the existing LGP formulation to support segmented optimization, which we term \textbf{Mini-LGP}:

\begin{equation}
\begin{aligned}
& \hspace{2mm} \min_{x, s_{1:K_\psi}, a_{1:K_\psi}, K_\psi} \hspace{2mm}\sum_{\psi=1}^{\Psi}\sum_{k=1}^{K_\psi}\int_{T_{k-1}}^{T_{k}} c(\bar{x}(t), s_k(t)) \, dt \\
& \hspace{1mm}\text{s.t.}\hspace{1mm} \quad \bar{x}(0) = 
\begin{cases}
    
    x_0& \text{if } \psi= 1\\
    x_{T_{K_{\psi -1}}}& \text{otherwise}
\end{cases}\\
& \forall k \in \{1, \ldots, K_\psi\} :\hspace{1mm}h_{\text{switch}}(\bar{x}(t_k), a_k) = 0, \\
& \hspace{36mm} g_{\text{switch}}(\bar{x}(t_k), a_k) \leq 0, \\
& \hspace{36mm}a_k \in A_\psi(s_{k-1}), \\
& \hspace{36mm}s_k \in \text{succ}(s_{k-1}, a_k), \\
& \hspace{36mm}s_{K_\psi} \in S_{\text{goal}}(g_\psi).
\end{aligned}
\label{mini_lgp}
\end{equation}

The environment offers $m$ distinct auxiliary objects that can share responsibility with the robot. Consequently, the problem can be segmented into a set of sub-problems $\Psi$, where $|\Psi|=2m$. Of these, $m$ are dedicated to fulfilling the necessary conditions for using these objects (e.g., filling the pitcher, moving objects onto a tray), while the remaining $m$ focus on executing shared responsibilities. Each sub-problem is defined and optimized jointly through Mini-LGPs over a problem-specific space. The initial configuration $\bar{x}(0)$ represents the starting state for the first sub-problem or the terminal state from the preceding sub-problem. The set of available actions, $A_\psi$, and the set of sub-goal states, $S_{\text{goal}}(g_\psi)$, are specific to each sub-problem (e.g., the robot cannot fill the glasses using a tray).

If a solution to Equation \ref{mini_lgp} exists, our formulation successfully generates a task plan using IRS as a heuristic with relaxed constraints. Now, our formulation will tackle the complete motion problem, $P_{path}$, considering the $P_{path}$ constraints with already computed mode-switch sequences $P_{seq}$. The complete IRS algorithm is shown in \textbf{Algorithm~\ref{IRS_ALG}}.

Depending on the auxiliary objects, it also returns a set of sub-problems, $\Psi$. These sub-problems are generic to each auxiliary object and require no further attention. We are not concerned with how to use the tray or pitcher, but when to use them. Hence, their utilization is assumed to be already defined based on the environment.

If the initial and goal states, $s_0$ and $s_g$, satisfy the generated conditions, the Mini-LGP solver, \texttt{solveMini-LGP}, attempts to find a solution with respect to the sequence bound of MBTS with its relaxed formulation defined in Equation \ref{mini_lgp}. If a plan can be found within this bound, the full problem is solved using the already found mode-switch constraints with \texttt{solvePathWithSequence}, which is a minimal version of Equation \ref{lgp} where only the configuration space is considered, without any symbolic search. If the full problem can be solved, the resulting motion plan $P_{path}$ is returned, where the use of auxiliary objects in this plan increases the agent's performance.

If the conditions generated by ORS are not satisfied, indicating that using auxiliary objects in $s_0$ and $s_g$ results in a performance decrease, then the LGP solver, \texttt{solvePath}, formulated in Equation \ref{lgp}, is used to find a motion plan. 

\begin{algorithm}[h]
\begin{footnotesize}
\caption{Interpretable Responsibility Sharing as a Heuristic}
\begin{algorithmic}[1] 
    \State \textbf{Input:} Dataset $D$, initial state $s_0$, initial kin. configuration $x_0\in X$, goal state $s_g$
    \State \textbf{Output:} Motion plan $P_{path}$
    \Function{Interpretable Responsibility Sharing}{$D, s_0, x_0, s_g$}
        \State $SequenceFound \gets false$
        \State $PathFound \gets false$
        \State Set of sub-problems $\Psi$, set of conditions for RS $S_c$ $\gets$ ORS(D)
        \If{$s_0 \cup s_g \in \{y\in X | \forall C_i\in S_c, C_i(y)\}$}
            \Repeat
                \Repeat
                    \State $P_{seq}$ $\gets$ $\textit{solveMini-LGP}(s_0, x_0, s_g,\Psi)$
                    \If{$P_{seq}$}
                        \State $SequenceFound\gets true$
                    \EndIf
                \Until{$SequenceFound$}
                \State $P_{path}$ $\gets$ $\textit{solvePathWithSequence}(s_0, x_0, s_g,P_{seq})$
                \If{$P_{path}$}
                    \State $PathFound\gets true$
                \EndIf
            \Until{$PathFound$}
        \Else
            \Repeat
                \State $P_{path}$ $\gets$ $\textit{solvePath}(s_0, x_0, s_g)$
                \If{$P_{path}$}
                    \State $PathFound\gets true$
                \EndIf
            \Until{$PathFound$}
        \EndIf
        \State \Return $P_{path}$
    \EndFunction
\end{algorithmic}
\label{IRS_ALG}
\end{footnotesize}
\end{algorithm}

\subsection{Computational Complexity}
\label{sec:complexity}

In this section, we analyze the computational complexity of the proposed framework, focusing on the tractability of the training phase (ORS) and the inference efficiency of the planning phase (IRS).

We define the following parameters for our analysis: $n$ denotes the number of training instances; $R$ represents the number of candidate atomic rules generated by RRL/CARL; and $L_{\max}$ is the maximum rule length.
For the iterative search, $\beta$ is the repetition count and $K_{\max}$ is the maximum size of the final rule set.

\subsubsection{Training Complexity}

The training process consists of candidate generation followed by rule selection.

\textbf{1. Candidate Generation (RRL \& CARL):}
RRL training scales linearly with the number of instances and active edges, with a per-epoch cost of $\mathcal{O}(n(b + W))$. CARL complexity is determined by path sampling and attention mechanisms over the knowledge graph, scaling as $\tilde{\mathcal{O}}(S \cdot \ell \cdot H \cdot d_{\text{emb}})$. Both steps are performed once per iteration and remain strictly polynomial.

\textbf{2. Optimized Rule Synthesis (ORS):}
To identify the optimal rule set without incurring the exponential cost of exhaustive enumeration ($\mathcal{O}(R^{K_{\max}})$), ORS utilizes a \textit{bounded greedy forward selection} strategy. In each of the $\beta$ repetitions, the algorithm iteratively adds the single rule $r \in \mathcal{R}$ that maximizes the validation score.

The cost to evaluate one candidate rule on $n$ validation samples is $\mathcal{O}(n \cdot L_{\max})$. Since the algorithm scans at most $R$ candidates for up to $K_{\max}$ steps across $\beta$ repetitions, the total complexity is:
\begin{equation}
    \mathcal{T}_{\text{ORS}} = \mathcal{O}(\beta \cdot K_{\max} \cdot R \cdot n \cdot L_{\max})
\end{equation}
This complexity is \textbf{linear} with respect to the candidate pool size $R$ and dataset size $n$. We constrain $L_{\max} \in \{3,4,5\}$, which ensures computational efficiency while enforcing TAMP-level interpretability, where rules remain concise yet descriptive enough to capture task context.

\subsubsection{Planning and Inference Efficiency}

It is crucial to note that the rule synthesis cost ($\mathcal{T}_{\text{ORS}}$) is a one-time offline expense. During online planning (inference), the agent merely evaluates the pre-generated rule set $S_c$.

We compare this online runtime against the Uninformed Search baseline used in our experiments. Uninformed Task and Motion Planning typically explores a state space with branching factor $b_s$ to depth $T$, yielding a complexity of $\mathcal{O}(b_s^T)$. By utilizing IRS, the task is decomposed into sub-problems. If IRS identifies a valid sub-goal (e.g., \textit{place on tray}), the effective search depth for the detailed motion planner is reduced to the sub-goal horizon $T_{sub} \ll T$, or the branching factor is pruned by focusing only on relevant auxiliary objects. Thus, IRS effectively reduces the exponent of the search complexity.

\section{Experiments}
This section primarily validates the effectiveness of IRS as a heuristic for Task and Motion Planning and the interpretability of the underlying decision-making mechanism, ORS, through several aspects. First, we compare our method's performance to other baselines in completing various household tasks, evaluating both the robot's effectiveness in minimizing displacement (representing the amount of effort) and the decision-making ability of ORS in determining the optimal time to share responsibility with auxiliary objects. Second, we compare human decision-making with ORS by observing how humans use auxiliary objects while completing a representative set of household tasks that ORS has also been tested on, analyzing the correspondence in object utilization to validate the human-centric environmental bias leveraged by IRS. Third, we present ablation studies on how ORS's dynamic integration behaves under the interpretability and accuracy trade-off.
\subsection{Tasks, Baselines and Evaluation Metrics}
\subsubsection{Tasks}
\begin{figure}[h]
  \centering
  \begin{subfigure}[b]{0.24\linewidth}
    \centering
    \includegraphics[width=\linewidth]{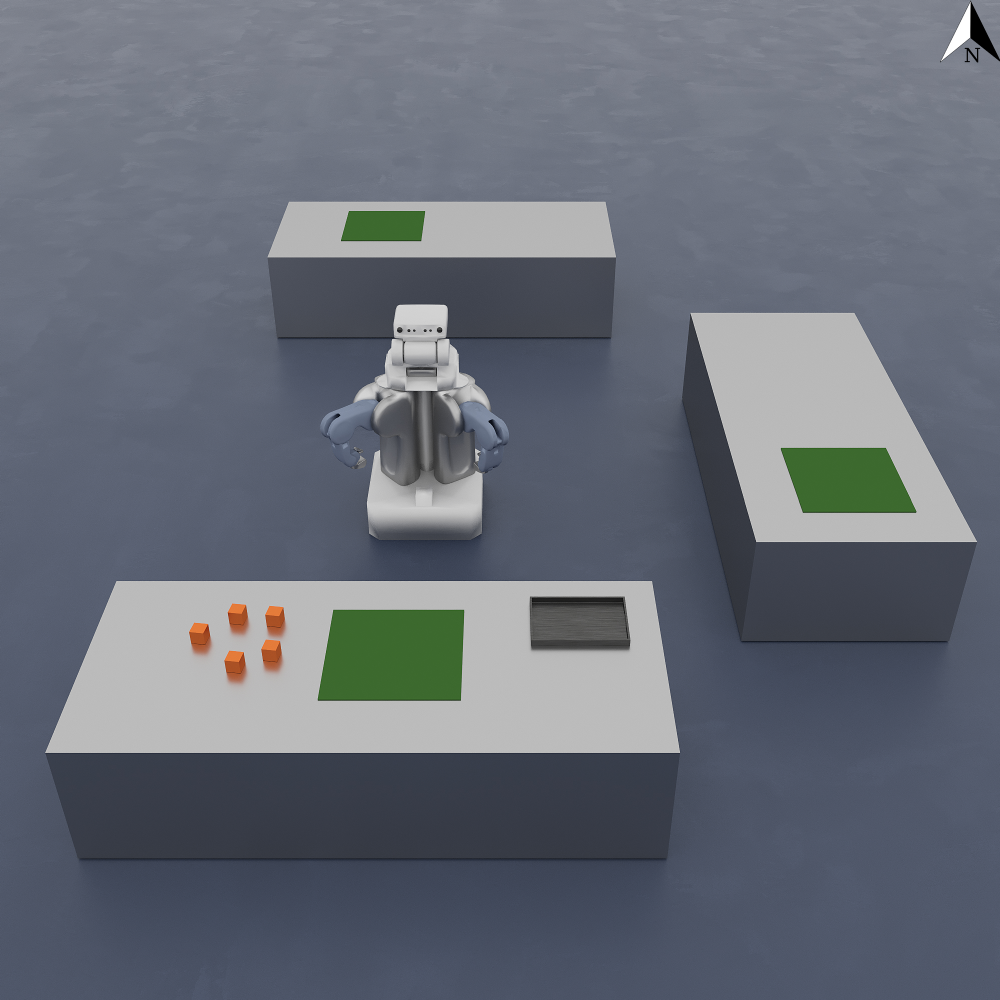}
    \caption{Serving Task}
    \label{fig:serving}
  \end{subfigure}
  \hfill
  \begin{subfigure}[b]{0.24\linewidth}
    \centering
    \includegraphics[width=\linewidth]{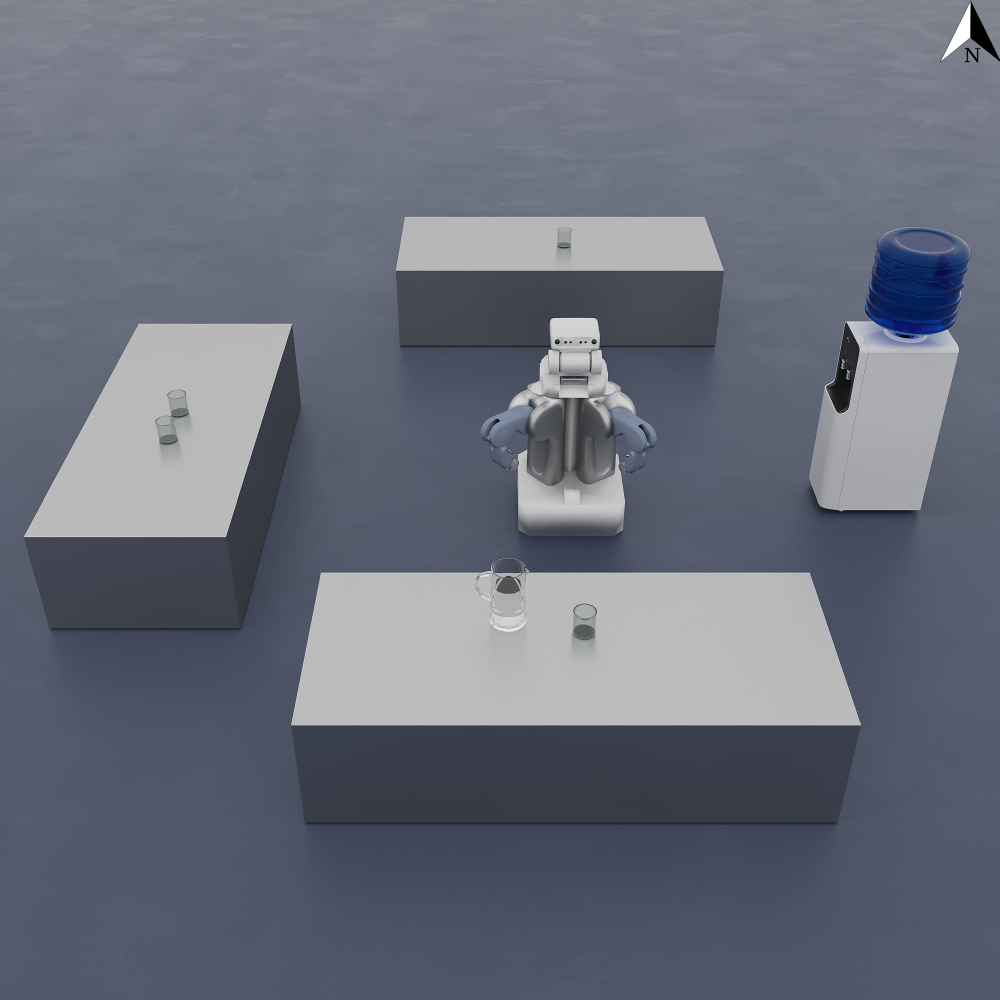}
    \caption{Pouring Task}
    \label{fig:pouring}
  \end{subfigure}
  \hfill
  \begin{subfigure}[b]{0.24\linewidth}
    \centering
    \includegraphics[width=\linewidth]{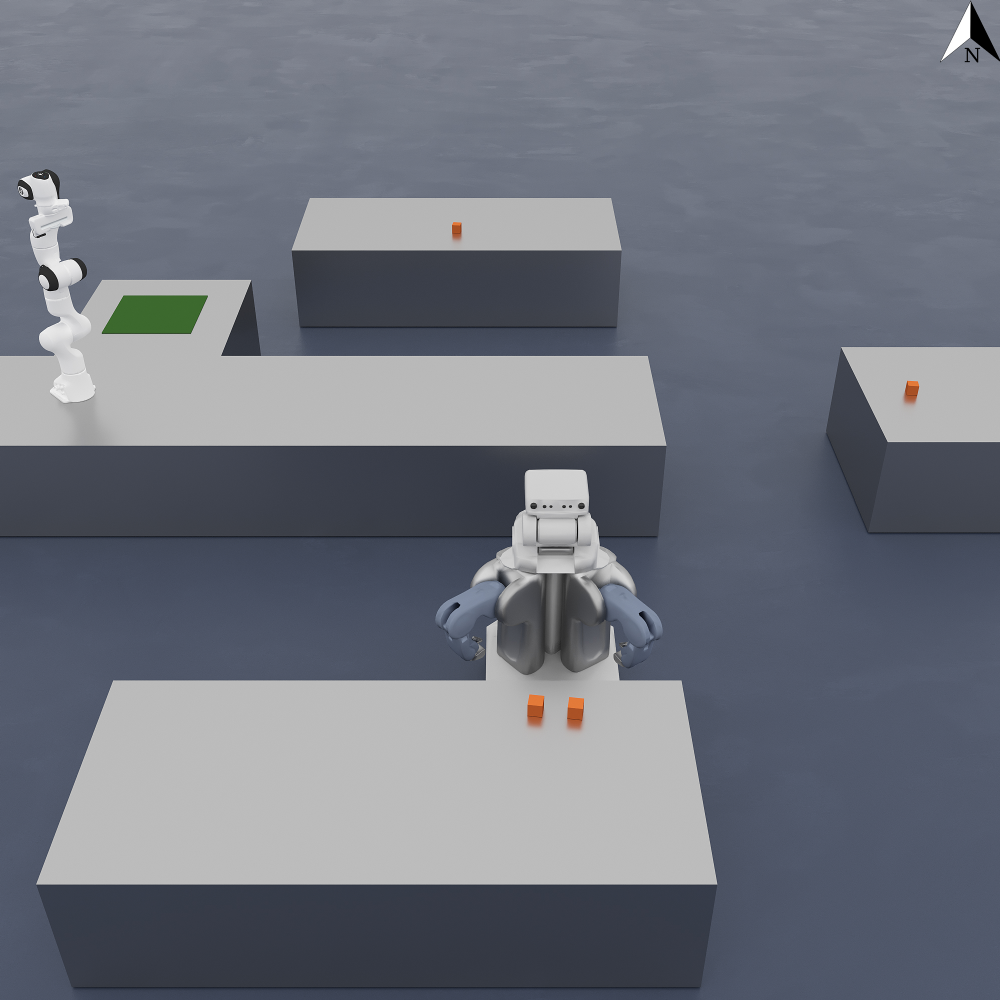}
    \caption{Handover Task}
    \label{fig:handover}
  \end{subfigure}
  \hfill
  \begin{subfigure}[b]{0.24\linewidth}
    \centering
    \includegraphics[width=\linewidth]{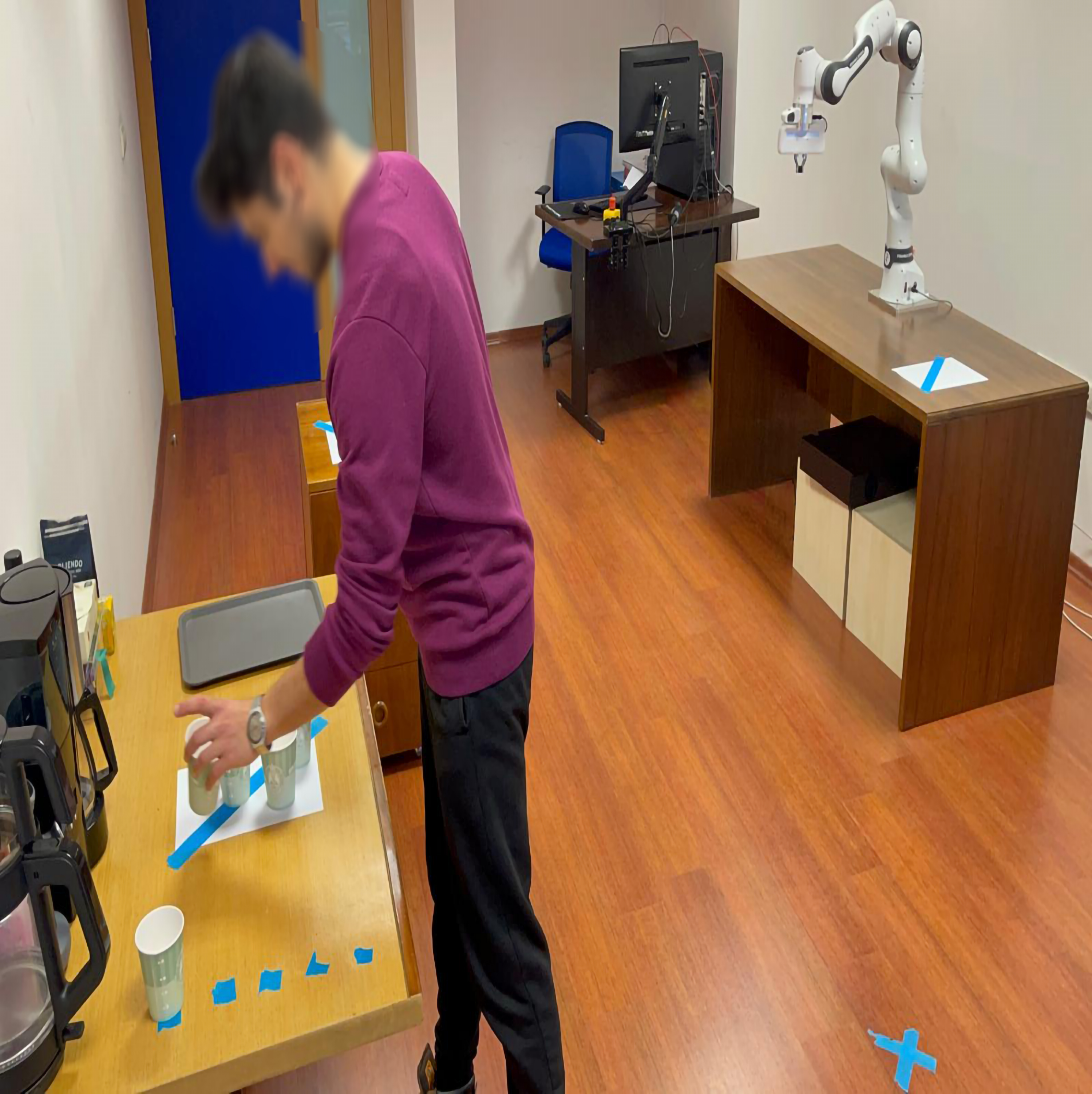}
    \caption{Human Experiments}
    \label{fig:human}
  \end{subfigure}
  
  \caption{\textbf{Experimental Environments and Task Configurations.} (a) \textbf{Serving Task:} A mobile PR2 robot transports objects from the counter to various dining tables, deciding whether to use a tray. (b) \textbf{Pouring Task:} The robot distributes water to glasses, choosing between individual filling or using a pitcher. (c) \textbf{Handover Task:} A multi-robot scenario where the mobile agent transfers objects to a stationary manipulator. (d) \textbf{Human Experiments:} A real-world setup replicating the Serving task to validate the correlation between human behavior and IRS decision-making. In simulation images, a \textit{North} sign provides spatial reference.}
  \label{fig:experiments}
\end{figure}
We evaluate IRS as a planning heuristic for task and motion planning in three distinct household tasks: serving, pouring, and handover. All of these tasks are modeled in simulation environments as replicas of real robotic tasks. 
To infer the usage of auxiliary objects for responsibility sharing with a single model, we utilized an overlapping set of descriptions for the environments, such as $table_{position}$ and $obj_{index}$. Additionally, since one of our main motivations in this paper is leveraging the human-centric environmental bias in sequential decision-making and manipulation problems through the concept of responsibility sharing, we wanted to investigate whether humans also utilize auxiliary objects when they provide a benefit. To do this, we structured a miniature but representative set of experiments based on the serving task and compared human behavior with IRS.

\par{\textbf{Serving:}}
In the serving task, a variable number of objects need to be carried from their initial position to their final destination. The agent can use a tray as an auxiliary object to share responsibility in completing the task. The setup includes a mobile PR2 robot with one active end effector and three tables—a counter, a kitchen table, and a dining table positioned to the south, east, and north—representing a typical kitchen scenario. Objects to be carried are always positioned on the counter, but the location of the tray and the final position (which can be one of the three tables) vary in each scenario, changing the effective distance between the initial and goal positions. This task aims to demonstrate how IRS will act in different arrangement scenarios (\textbf{Figure \ref{fig:serving}}).

\par{\textbf{Pouring:}}
In the pouring task, a variable number of glasses need to be filled by a mobile PR2 agent. The glasses can either be filled directly from a water source, which the agent should place them on, or the agent can fill a pitcher and then distribute the water to each glass. Similar to the serving task, there are three tables positioned at the north, south, and west. The glasses and pitcher can be on any table, but the water source is fixed at the east of the environment. We aim to investigate the effectiveness of IRS in various distribution scenarios in this task (\textbf{Figure \ref{fig:pouring}}).

\par{\textbf{Handover:}}
Similar to the serving task, the handover task requires objects to be carried from their starting position to their final position. However, the environment is designed as a U-shaped maze with table placements. In this complex structure, an additional stationary Franka robot is included so that the mobile PR2 agent can deliver the objects to the stationary robot, which then places the objects in the final position, rather than the PR2 doing it itself. Objects can initially be positioned on any table at the north, south, or east, and they need to be carried to a fixed final position. This task demonstrates how an embodied agent can also be viewed as an auxiliary object that shares responsibility with the primary agent. It expands our experiments to a multi-robot setting and reinforces the concept of responsibility sharing, as humans typically share their responsibilities with other agents (\textbf{Figure \ref{fig:handover}}).

\par{\textbf{Human Experiments:}}
The development of IRS is notably influenced by intuitive human thinking, making human experiments a crucial aspect of our study to observe human-centric environmental bias. For these experiments, we replicated a portion of the serving task using a varying number of objects (mugs) and a single tray initially positioned on the counter. We asked six human subjects to carry the mugs to their final positions based solely on intuition, without specific instructions regarding the use of auxiliary objects. The primary objective of these experiments was to provide empirical grounding for our core hypothesis: that the planning logic modeled in our framework reflects genuine human strategic intuition. To observe this human-centric environmental bias in a condensed and observable form, we designed the experiments to isolate the logical relationship between transport distance and object count from confounding physical variables. We evaluated the alignment by comparing human decisions in these physical trials against the ORS model's predictions for semantically equivalent scenarios within the simulation (\textbf{Figure \ref{fig:human}}).

\subsubsection{Baselines}
We compare IRS with several baselines to evaluate its effectiveness. First, we compare it with LGP ~\cite{lgp, toussaint2018differentiable}, which does not prioritize using auxiliary objects, and a baseline agent that always prioritizes using auxiliary objects, to show that responsibility sharing depends on the environmental setting and that IRS can improve the existing TAMP formulation. Then, we compare ORS's decision-making performance with well-established baselines, including Artificial Neural Network (ANN) ~\cite{Rosenblatt1958ThePA}, XGBoost ~\cite{10.1145/2939672.2939785}, Support Vector Machines (SVM) ~\cite{hearst1998support}, Logistic Regression ~\cite{cox1958regression}, and Decision Tree ~\cite{quinlan1986induction}. Among these baselines, ANN and Decision Tree represent two opposite ends of the interpretability spectrum, providing a range for comparison. Additionally, the performance of RRL ~\cite{wang2021scalable} and CARL ~\cite{he2023correlation} is investigated, enabling comparison with interpretable state-of-the-art methods.
\subsubsection{Evaluation Strategy}
For baseline comparison and ablation studies, we constructed a dataset using the process explained in Section \ref{dataset} and performed randomized split testing, repeated across 5 different random seeds, including all tasks collectively. In all experiments, we fix the ORS repetition parameter to $\beta = 10$, corresponding to the upper end of the stable range identified in our preliminary tuning (we observed that $\beta \in [5,10]$ yields similar validation performance, while smaller values lead to less stable rule sets across runs).

We then calculated the following evaluation metrics on test scenarios: Effort, Accuracy, Precision, Recall, F1-Score, Confidence, and Interpretability. Effort is measured by the total unit displacements of the robot's joints in the environment required to complete the task. Accuracy, Precision, Recall, and F1-Score metrics were used to measure decision-making performance in terms of using IRS as a heuristic. Confidence and Interpretability are used to assess the quality and explainability of the proposed model and existing interpretable methods. Findings are reported as the mean and standard deviation across the folds.

To specifically assess the reliability of individual symbolic rules generated by the models, we define a \textbf{Rule Confidence} score, $\mathrm{Conf}(r)$. This score estimates the conditional probability that a rule's prediction is correct given that its antecedent holds. Formally, let $D_{\text{val}}$ denote the validation split, and let $\varphi_r(x_i)\in\{0,1\}$ indicate whether the antecedent of rule $r$ is satisfied on instance $x_i$. Let $h_r(x_i)$ be the label predicted by $r$, and $y_i$ the ground-truth label. We calculate confidence as the empirical precision on the covered validation instances:
\begin{equation}
\mathrm{Conf}(r) = \frac{\sum_{i \in D_{\mathrm{val}}} \mathbb{I}\big[\varphi_r(x_i)=1 \wedge h_r(x_i)=y_i\big]}
     {\sum_{i \in D_{\mathrm{val}}} \mathbb{I}\big[\varphi_r(x_i)=1\big] + \varepsilon},
\label{eq:rule_confidence}
\end{equation}
where $\varepsilon>0$ is a small constant for numerical stability. It is important to note that $\mathrm{Conf}(r)$ evaluates the rule at the aggregate level (the posterior reliability). It does not assign weights to individual conditions within the antecedent; rather, the relative importance of specific features and conditions is encoded intrinsically during the rule generation phase, via multi-head attention scores in CARL~\cite{he2023correlation} and Gradient Grafting weights in RRL~\cite{wang2021scalable}. Therefore, $\mathrm{Conf}(r)$ serves as a post-hoc measure of the resulting rule's quality rather than a feature importance metric.
\subsection{Data Preprocessing}
The RRL model utilizes structure-based data, which emphasizes the relationships and hierarchies between different elements. Conversely, the CARL model uses a knowledge graph-based dataset, where the data is structured as a network of entities connected by relations. To integrate the rules derived from this structured data into both RRL and CARL models, we convert the data into a binary conditions suitable for RRL, while for CARL, the data is formatted as a knowledge graph.
\subsection{Results}
\subsubsection{Performance Comparison}
\begin{table}[ht]
    \begin{tiny}
    \centering
    \begin{tabular}{cccccc}
        \toprule
         & Accuracy $\uparrow$& Precision $\uparrow$& Recall $\uparrow$& F1 Score $\uparrow$& Effort $\downarrow$\\
        \midrule
        LGP & - & - & - & - & $13.47 \pm 6.21$ \\
        Control & - & - & - & - & $13.37 \pm 5.83$ \\
        Decision Tree & $0.705 \pm 0.054$ & $0.717 \pm 0.046$ & $0.702 \pm 0.056$ & $0.696 \pm 0.061$ & $12.06 \pm 5.69$\\
        Logistic Regression & $0.720 \pm 0.089$ & $0.745 \pm 0.080$ & $0.726 \pm 0.084$ & $0.715 \pm 0.091$ & $11.80 \pm 5.65$\\
        SVM & $0.760 \pm 0.093$ & $0.781 \pm 0.079$ & $0.764 \pm 0.084$ & $0.756 \pm 0.092$ & $11.71 \pm 5.52$\\
        XGBoost & $0.755 \pm 0.102$ & $0.774 \pm 0.092$ & $0.752\pm 0.098$ & $0.747 \pm 0.104$ & $11.97 \pm 5.71$\\
        ANN &  $0.805 \pm 0.069$ & $0.817 \pm 0.062$ & $0.802 \pm 0.070$ & $0.800 \pm 0.071$ & $11.49 \pm 5.35$ \\
        RRL & $0.781 \pm 0.077$ & $0.780 \pm 0.062$ & $0.784 \pm 0.070$ & $0.781 \pm 0.072$ & $11.49 \pm 4.80$ \\
        CARL & $0.950 \pm 0.056$ & \textbf{0.961 $\pm$ 0.066} & $0.970 \pm 0.068$ & $0.961 \pm 0.059$ & $11.35 \pm 4.64$\\
        \textbf{ORS (ours)} & \textbf{0.963 $\pm$ 0.004} & $0.954 \pm 0.004$ & \textbf{0.980 $\pm$ 0.002} & \textbf{0.980$\pm$0.003} & \textbf{11.05 $\pm$ 4.79}\\
        \bottomrule
    \end{tabular}
    \caption{\textbf{Quantitative Performance Comparison.} Evaluation of IRS (using ORS) against baselines across Serving, Pouring, and Handover tasks. Metrics include decision-making performance (Accuracy, Precision, Recall, F1) and physical task efficiency (Effort, measured in joint displacement). Values represent mean $\pm$ standard deviation across 5-fold cross-validation. \textbf{Bold} indicates the best performance. Note that LGP and Control are fixed strategies (standard search vs. forced auxiliary usage) and do not have classification metrics ('-').}
    \label{acc_results}
    \end{tiny}
\end{table}

We compare the performance of IRS as a heuristic for TAMP with a well-established formulation, LGP. To validate that Responsibility Sharing adapts to the environment and changes depending on the initial and goal states, we also implemented a control baseline that always utilizes auxiliary objects available in the environment. 

In the five-fold cross-validation test environments, IRS achieved a minimum effort of $11.05 \pm 4.79$, while LGP achieved $13.47\pm6.21$ and the control achieved $13.37\pm5.83$ (\textbf{Table~\ref{acc_results}}). LGP performs better when there are fewer target objects (1 to 2) in the environment, while the control performs better with an increased number of target objects (greater than 2), showing the exponential improvement of responsibility sharing. Effort-based evaluations validate that IRS improves the agent's effectiveness due to its dynamic heuristic selection. Additionally, the difference between the control and LGP suggests that always using auxiliary objects can improve performance but may not be the optimal behavior for the agent.

It is worth noting that the standard deviation for the Effort metric is relatively high across all methods. This is primarily attributed to the diversity of the test scenarios across different experimental settings, which feature a varying number of target objects (ranging from 1 to 5) and differing distances between initial and goal configurations. Consequently, the baseline physical effort required varies significantly between task instances. However, IRS achieves the lowest standard deviation ($4.79$) compared to LGP ($6.21$) and the Control ($5.83$), indicating that our method not only reduces the mean effort but also provides more consistent performance across diverse task complexities.

Furthermore, we conducted extensive evaluations on the decision-making capacity of our proposed method, ORS. The quantitative results are shown in Table \ref{acc_results}. ORS outperforms all baselines across three tasks, achieving significant decision-making accuracy, which is reflected in its performance as it requires the minimum effort to complete these tasks. 
The quantitative results reveal a clear performance hierarchy. The ANN, RRL, and CARL models show notable performance, with ANN achieving an accuracy of $80.5 \pm 6.9\%$ and RRL achieving $78.1 \pm 7.7\%$. Among the baselines, CARL demonstrates superior predictive power with an accuracy of $95.0 \pm 5.6\%$, indicating that its relational rules are the primary drivers of performance.

However, our proposed method, ORS, significantly outperforms all baselines, achieving the highest accuracy of $96.3 \pm 0.4\%$. This enhancement is due to the synergy between the learners: ORS synthesizes the high-performing relational rules from the joint rule pool with the structural, interpretable logic. By employing a bounded greedy search to optimize a Balance Score, ORS filters out unstable rules, reducing the standard deviation by an order of magnitude (from $\pm 5.6\%$ in CARL to $\pm 0.4\%$ in ORS). Thus, ORS leverages CARL for raw predictive power while relying on RRL's structural constraints to ensure stability and interpretability.

\par{\textbf{Error Analysis and Geometric Variability:}}
While ORS achieves high accuracy ($\approx 96.3\%$), the majority of the remaining misclassifications highlight the distinction between the continuous geometric costs used for ground-truth labeling and the discrete logical predicates used for inference. The CPG module generates labels based on the precise cost of executed motion plans; consequently, it effectively identifies scenarios where auxiliary objects are present but geometrically suboptimal (e.g., located at the far edge of the workspace) as negative. In contrast, the ORS model operates on high-level logical features (e.g., $tray\_on\_table$) that abstract away specific metric coordinates. As a result, the model learns generalized semantic rules, such as utilizing a tray for multiple objects, which may occasionally contradict the specific geometric ground truth in boundary cases. These discrepancies represent a trade-off inherent to the framework, prioritizing the learning of robust, interpretable strategies over the overfitting of specific geometric initializations.
\subsubsection{Confidence Comparison}
\begin{table}[ht]
    \centering
    \resizebox{\textwidth}{!}{
    \begin{tabular}{ccc}
        \toprule
         Models & Sample Rules & Confidence \\
        \midrule
        \multirow{2}{*}{RRL} & $negative$ $\longleftarrow$ $obj2$ $on$ $table_{east}$ $|$ $tray$ $on$ $table_{north}$ $|$ $jug$ $on$ $table_{west}$ & 0.4188 \\ 
        & $positive$ $\longleftarrow$ $obj0$ $on$ $table_{south}$ $|$ $obj1$ $on$ $table_{south}$ $|$ $\neg goal$ $at$ $table_{south}$ $|$ $\neg goal$ $at$ $table_{east}$ $|$ $\neg action_{fill}$ & 0.5938\\
        \multirow{2}{*}{CARL} & $negative$ $\longleftarrow$ $obj4$ $on$ $table_{south}$ $|$ $\neg goal$ $at$ $table_{handover}$ $|$ $\neg obj0$ $on$ $table_{south}$ & 0.6244 \\ 
        & $positive$ $\longleftarrow$ $obj0$ $on$ $table_{east}$ $\&$ $\neg obj0$ $on$ $table_{south}$ & 0.9408\\
        \multirow{2}{*}{ORS} & $negative$ $\longleftarrow$ $obj0$ $on$ $table_{south}$ $\&$ $obj1$ $on$ $table_{south}$ $\&$ $obj2$ $on$ $table_{south}$ $\&$ $obj3$ $on$ $table_{south}$ $\&$ $\neg goal$ $at$ $table_{south}$ & 0.7705 \\ 
        & $positive$ $\longleftarrow$ $obj0$ $on$ $table_{south}$ $\&$ $is$ $helper$ $exist$ $\&$ $\neg goal$ $at$ $table_{south}$ $\&$ $\neg goal$ $at$ $table_{east}$ $\&$ $\neg action_{fill}$ $\&$ $\neg action_{pour}$ & 0.9801\\
        \bottomrule
    \end{tabular}%
    }
    \caption{\textbf{Sample Rules and Confidence Scores.} A comparison of extracted rules from RRL, CARL, and ORS. The \textit{Sample Rules} column shows the logical preconditions that trigger a prediction. \textit{Positive} rules indicate that utilizing an auxiliary object reduces physical effort (Responsibility Sharing), while \textit{Negative} rules advise against it. ORS synthesizes rules with higher specificity (combining multiple spatial and action predicates via conjunctions $\&$), yielding significantly higher confidence scores compared to the baselines. This illustrates ORS's ability to capture complex, context-dependent dependencies.}
    \label{confidence_rule_results}
\end{table}
\textbf{Table \ref{confidence_rule_results}} provides a comparative analysis of rule-based confidence scores across three models: RRL, CARL, and ORS. Each model employs distinct rules for decision-making in task and motion planning. The RRL model categorizes its rules into negative and positive based on object placements and interactions, while the CARL model uses conditions involving object interactions and goal locations, resulting in varying confidence levels for negative and positive rules. 

The ORS model, aimed at optimized rule synthesis, augment conditions by combining object locations and actions, yielding high confidence scores, especially for positive scenarios. This table highlights the complex interplay between spatial configurations and model predictions, showcasing the models' capabilities to synthesize and apply rules in intricate task environments. ORS was able to generate rules with the highest confidence and interpretability, involving a higher number of rule combinations. Qualitatively, it is also observed that despite having high confidence, CARL generated less meaningful rules for the negative class. This is because there is no instance of the goal position being at the handover table, which adds no additional information to the rule combinations.
\subsubsection{Responsibility Sharing in Humans}
\begin{table}[h!]
    \centering
    \begin{tabular}{cccc}
        \toprule
        \# of Mugs & Close & Medium & Far \\
        \midrule
        1 & 100\% & 100\% & 100\% \\
        2 & 100\% & 83.33\% & 50\% \\
        3 & 100\% & 100\% & 100\% \\
        4 & 100\% & 100\% & 100\% \\
        5 & 100\% & 100\% & 100\% \\
        \bottomrule
    \end{tabular}
    \caption{\label{human_performance}\textbf{Alignment between Human and ORS Decision-Making.} The table presents the percentage of human participants ($N=6$) whose decision to utilize the tray matched the decision predicted by the ORS model. The evaluation covers scenarios with varying object counts (1--5 mugs) and transport distances (Close, Medium, Far) between the counter and the target table. A value of 100\% indicates perfect consensus between human intuition and the learned logical rule. The divergence in the 2-mug/Far case highlights a decision boundary where the utility of the tray becomes ambiguous for humans, whereas ORS maintains a consistent strategy.}
\end{table}
We conducted human experiments to compare ORS with human subjects regarding their choices in utilizing auxiliary objects and sharing task responsibilities. These experiments focused on the serving task under constrained scenarios, where a varying number of objects and a tray were initially located on a table representing the counter. The goal positions varied among three options to create differences in both the distance (Close, Medium, Far) and the number of objects to be carried. \textbf{Table~\ref{human_performance}} compares human behavior with ORS's object utilization, providing insights into Responsibility Sharing (RS) in humans. The percentages indicate the proportion of participants whose tray usage aligned with ORS's. The experimental setup in the simulation closely matched real-world scenarios in terms of both position and distance, enabling us to observe implicit human behavior regarding the use of auxiliary objects. Reflecting on the findings in Table~\ref{human_performance}, we emphasize that in 13 out of 15 experimental settings, ORS's behavior matched human actions in using the tray. 

This highlights two points: the presence of Human-Centric Environmental Bias in human decision-making, and the high matching ratio indicating that ORS effectively captures this intrinsic environmental structure, even though imitating human behavior was not the primary objective.
\subsection{Ablation Studies on Interpretability and Accuracy Trade-Off}
\begin{figure}[h]
    \centering
    \includegraphics[scale=.50]{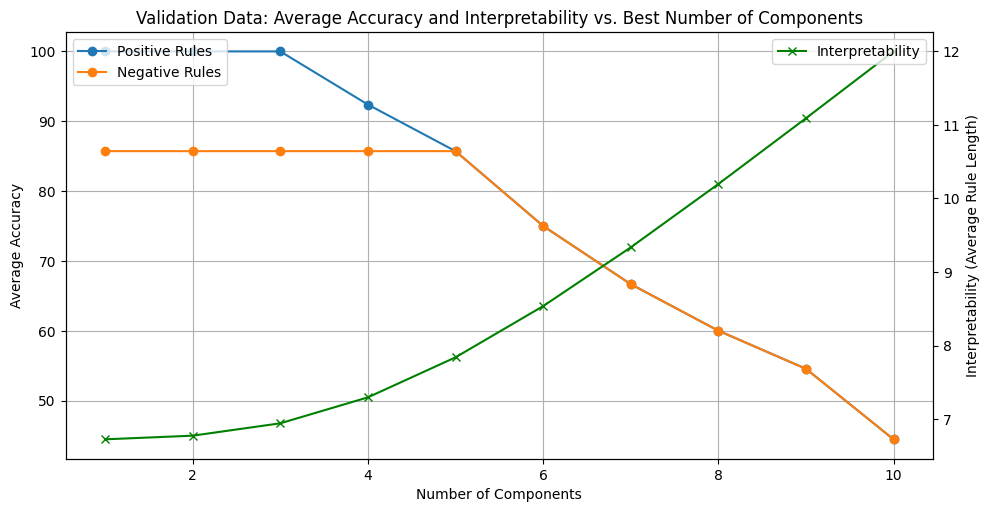}
    \caption{\textbf{Impact of Rule Complexity on Model Performance.} The graph illustrates the trade-off between predictive accuracy and interpretability within the ORS framework. The x-axis represents the number of rule components selected. The left y-axis tracks the Average Accuracy for both positive and negative rules, while the right y-axis measures Interpretability (defined as the Average Rule Length). As the number of components increases, the rules become more descriptive (higher interpretability) but less generalizable, leading to a drop in accuracy. ORS optimizes this trade-off by selecting a rule set that balances high accuracy with sufficient causal descriptiveness.}
    \label{fig:trade}
\end{figure}
As interpretability increases—indicated by the average rule length in the model—the model becomes more transparent and easier to understand. However, this often results in a decrease in accuracy. This reduction occurs because a higher number of components (rules) can make the model dense, potentially introducing complexity that does not effectively capture the underlying patterns, as seen in the decreasing trend of average accuracy for positive and negative rules (\textbf{Figure \ref{fig:trade}}). Conversely, when fewer rules are used, the model becomes sparse and may focus on capturing the most significant patterns, leading to higher accuracy but at the cost of reduced interpretability. This balance highlights the challenges in achieving both high accuracy and high interpretability in model development, emphasizing the need to carefully consider the impact of rule complexity and sparsity on model performance.

\section{Discussions}
\subsection{Human-Centric Environmental Bias and Responsibility Sharing}
IRS is a promising framework that leverages human-centric environmental bias to enhance the classical TAMP formulation. As shown in Table~\ref{acc_results}, its application to various household tasks has been notably successful. IRS outperformed classical TAMP approaches, reinforcing our initial hypothesis that humans are a common factor in both environment construction and task design; therefore, an agent should utilize the auxiliary objects available in the environment. Furthermore, we investigated the practicality and effectiveness of always using an object with our control agent. The results show that solutions to sequential decision-making and manipulation problems are highly dependent on the initial configuration. This underscores the need for an adaptive use of the responsibility-sharing heuristic, which we address with ORS. 
Finally, our human experiments demonstrated a consistent alignment between the concept of responsibility sharing and human decision-making. By comparing human decisions in physical trials against the ORS model's predictions for semantically equivalent scenarios within the simulation, we verified that the choice to utilize auxiliary objects was driven by a strong, intuitive signal inherent in the environment. These results validate the existence of Human-Centric Environmental Bias, showing that humans naturally leverage these environmental structures (e.g., trays) to optimize effort. By capturing this behavior through interpretable rules rather than black-box approximations, ORS provides a transparent mechanism that aligns robotic planning with the logical heuristics used by humans.

\subsection{Rationale for Comparing Decision Trees with ORS}
In the evaluation of rule-based decision-making systems, comparing decision trees and the ORS model is essential for several key reasons. This comparison highlights the strengths and limitations of each approach. Decision trees are known for their simplicity and high interpretability, making them ideal for straightforward hierarchical decision scenarios. In contrast, our ORS model excels in handling complex relational dynamics and rule sequences due to the integration of advanced rule learning techniques like CARL and RRL~\cite{Garcez2023}. In sequential decision-making and manipulation problems, complex relational dynamics become increasingly relevant, highlighting the need for an advanced approach like ORS, as supported by the reduced overall effort required by the model to complete tasks.
\subsection{Accuracy Interpretability Trade-off}
In selecting baseline models, we evaluated a range of methods, including state-of-the-art interpretable methods such as RRL and CARL, as well as several models ranging from decision trees to ANNs. The motivation behind this selection was to create a spectrum of models representing different levels of interpretability. Our experiments showed that, with the exception of CARL and ORS, model performance decreased as interpretability increased. This finding was further supported by our ablation studies, which revealed a clear trade-off between interpretability and performance. For CARL, the divergence is likely due to its integration of deeper semantic knowledge in the generated rules. In the case of ORS, it is attributed to its optimized rule synthesis mechanism, as indicated by the balance score.
\section{Limitations}
IRS proposes a novel approach to improve the effectiveness and interpretability of existing TAMP formulations through the concept of responsibility sharing. However, this approach entails specific limitations, particularly regarding physical realism.

\textbf{Physical Capacity and Constraints:} Currently, ORS operates under a discrete state representation that assumes auxiliary objects have sufficient capacity to handle assigned tasks (e.g., a tray can carry all target objects). We acknowledge that this is a critical assumption. However, the modular design of IRS provides a \textbf{concrete path} to relax this assumption in future work:
\begin{enumerate}
    \item \textbf{Solver-Level Integration:} The underlying \textit{Mini-LGP} solver (Eq.~\ref{mini_lgp}) can enforce capacity constraints during the sub-problem optimization phase. If a heuristic-suggested plan violates physical limits (e.g., objects do not fit on the tray), the solver will deem the sub-goal infeasible, forcing the planner to fall back to standard search or split the task into multiple trips.
    \item \textbf{Feature-Level Integration:} To make the decision-making \textit{proactive} rather than reactive, continuous affordance properties (e.g., $V_{\textrm{objects}}$ vs.~$C_{\textrm{tray}}$, for volume and capacity, respectively) can be integrated into the ORS feature space $\Phi$. This would allow the rule synthesizer to learn capacity-aware conditions (e.g., \textit{Use Tray IF Count $>$ 2 AND FitsOnTray}), bridging the gap between logical heuristics and physical constraints.
\end{enumerate}

\textbf{Partial Observability:} Our model currently assumes fully observable domains. Extending IRS to partially observable settings would require integrating a scene graph generation module to dynamically populate the first-order logic state representation $S$ from raw sensory data. Provided that the initial state can be inferred, the underlying reasoning mechanism of IRS remains applicable.

\textbf{Cognitive Complexity:} Future studies will explore complex, cognitively demanding tasks where human intuition may diverge from optimal planning. We assume that such divergences are not failures of the model, but rather indicators of tasks that exceed human cognitive efficiency. Identifying scenarios where automated planning outperforms human baselines could reveal critical opportunities for automation transformation in domestic environments.
\section{Conclusion}
We present IRS, a novel heuristic for task and motion planning that leverages human-centric environmental bias by incorporating responsibility sharing with auxiliary objects through its interpretable decision-making mechanism, ORS. ORS is trained with a dataset created by CPG, containing samples that indicate whether using auxiliary objects will improve the agent's performance. The optimal integration of rules by ORS ensures transparency and applicability, making it an effective decision-making mechanism for IRS heuristic adaptation. Experiments in household tasks demonstrate that IRS significantly improves agents' effectiveness compared to previous TAMP formulations, and ORS outperforms existing baselines, achieving the highest accuracy and interpretability.
\section*{CRediT authorship contribution statement}
\textbf{Arda Sarp Yenicesu:} Conceptualization, Data curation, Formal analysis, Investigation, Methodology, Visualization, Writing – original draft.  \textbf{Sepehr Nourmohammadi:} Conceptualization, Formal analysis, Investigation, Methodology, Writing – original draft. \textbf{Berk Cicek:} Formal analysis, Validation, Writing – original draft.  \textbf{Ozgur S. Oguz:} Conceptualization, Funding acquisition, Project administration, Supervision, Writing – review and editing.
\section*{Declaration of competing interest}
The authors declare that they have no known competing financial
interests or personal relationships that could have appeared to influence
the work reported in this paper.
\section*{Data Availability}
Data will be made available on request.
\section*{Ethics Approval}
This work involved human subjects in its research. Approval for all ethical and experimental procedures and protocols was granted by the Institutional Review Board of Bilkent University under Application No. 2024\_02\_12\_01, and the study was conducted in accordance with the Declaration of Helsinki.
\section*{Participant Consent}
This work involved human subjects in its research. Participants were recruited on a voluntary basis without any specific requirements, and informed consent was obtained from all participants.
\section*{Acknowledgements}
Funding: This work was supported by TUBITAK under 2232 program with project number 121C148 (“LiRA”). Any opinions, ﬁndings, and conclusions or recommendations expressed in this material are those of the authors and do not necessarily reﬂect the views of the TUBITAK. 

We would like to express our sincere gratitude to Begüm Başar for her invaluable contribution to the art design and visualization of the graphical abstract and experimental setup. We also extend our thanks to Dilruba S. Haliloglu and Kutay Demiray for their meticulous proofreading.
\bibliographystyle{elsarticle-num}
\bibliography{reference}

\end{document}